\documentclass[acmtog, authorversion]{acmart}
\acmSubmissionID{593}

\usepackage{booktabs} 
\usepackage{minted}
\usepackage[capitalize]{cleveref}
\usepackage{multirow}
\crefname{section}{Sec.}{Secs.}
\crefname{table}{Tab.}{Tabs.}
\usepackage[normalem]{ulem}
\useunder{\uline}{\ul}{}
\usepackage[table,xcdraw]{xcolor}

\definecolor{linkblue}{RGB}{0,102,204}
\hypersetup{urlcolor=linkblue}

\renewcommand\footnotetextcopyrightpermission[1]{}

\usepackage[normalem]{ulem}

\usepackage{subcaption}
\usepackage{graphicx}
\usepackage{hyperref}

\usepackage[ruled]{algorithm2e} 

\SetAlFnt{\small}
\SetAlCapFnt{\small}
\SetAlCapNameFnt{\small}
\SetAlCapHSkip{0pt}

\usepackage{xcolor}

\newcommand{\link}[1]{\textcolor{linkblue}{ #1}}

\newcommand{\wgn}{white Gaussian noise}
\newcommand{\Wgn}{White Gaussian noise}

\begin{document}
\title{Colorful-Noise: Training-Free Low-Frequency Noise Manipulation for Color-Based Conditional Image Generation}

\author{Nadav Z. Cohen}
\orcid{1234-5678-9012-3456}
\affiliation{%
 \institution{Reichman University}
 \country{Israel}}

\author{Ofir Abramovich}
\affiliation{%
 \institution{Reichman University}
 \country{Israel}
}

\author{Ariel Shamir}
\affiliation{%
\institution{Reichman University}
\country{Israel}}


\begin{abstract}
Text-to-image diffusion models generate images by gradually converting \wgn\ into a natural image. \Wgn\ is well suited for producing diverse outputs from a single text prompt due to its absence of structure. However, this very property limits control over, and predictability of, specific visual attributes, as the noise is not human-interpretable. In this work, we investigate the characteristics of the input noise in diffusion models. We show that, although all frequencies in \wgn\ have comparable statistical energy, low-frequency components primarily determine the image’s global structure and color composition, while high-frequency components control finer details. Building on this observation, we demonstrate that simple manipulations of the low-frequency noise using low-frequency image priors can effectively condition the generation process to reconstruct these low-frequency visual cues. This allows us to define a simple, training-free method with minimal overhead that steers overall image structure and color, while letting high-frequency components freely emerge as fine details, enabling variability across generated outputs.
\end{abstract}




\begin{teaserfigure}
  \includegraphics[width=\textwidth]{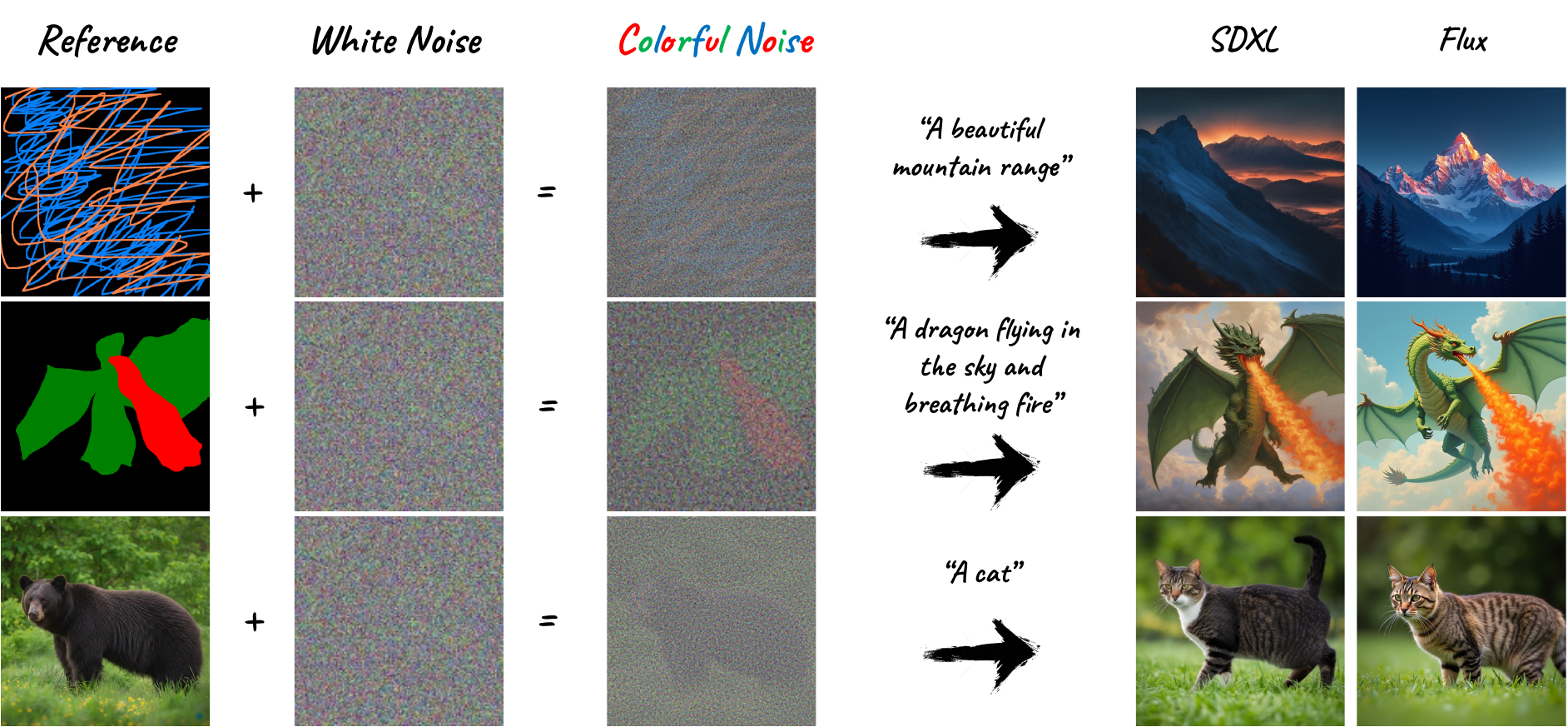}
  \caption{\textbf{Colorful-Noise.} Conditioning the low-frequency components of \wgn\ with structured color maps enables control over both image structure and color scheme, without requiring training or incurring any additional computational overhead (our method works in latent space but examples shown here are in pixel space for illustrative purposes).}
  \label{fig:teaser}
  \Description{A teaser for our paper}
\end{teaserfigure}

\maketitle
\pagestyle{plain}
\fancyfoot{}
    


\section{Introduction}
Latent Diffusion and flow-matching models are the current state of the art for visual content generation. To generate images, they learn to map \wgn\ latents to image latents by a gradual multi-step denoising process. The stochastic nature of \wgn\ is central to a model’s ability to produce diverse outputs, allowing a single text prompt to produce a wide range of visual concepts when paired with different noise samples. However, while this randomness enables diversity, it also imposes a fundamental limitation: the noise is not interpretable to humans, making it difficult to control or predict specific visual attributes of the generated image based on the noise alone.

Many recent works have explored methods for controlling and editing specific visual aspects of generated images. These aspects can be broadly categorized into \emph{content-control} -- governed by structural guidance (e.g., layout maps or sketches) or descriptive text prompts, and \emph{style-control} -- using reference images or stylistic text descriptions. To achieve control, existing approaches employ a range of techniques, including conditional prior modeling, LoRA fine-tuning~\cite{hu2021loralowrankadaptationlarge}, attention manipulation~\cite{attention_is_all_you_need}, and noise inversion~\cite{song2022denoisingdiffusionimplicitmodels}. In contrast, direct manipulation of the latent noise itself remains relatively underexplored, largely due to the fragility of diffusion models and the inherently unstructured nature of \wgn. 

In this work, we introduce Colorful-Noise, a method for conditioning diffusion noise latents with structured color maps, motivated by an analysis of the relationship between the noise input space and the generated image space. Although noise latents are not directly interpretable by humans, we observe that they encode different types of visual information across frequency bands: high frequencies primarily correspond to fine details and textures, whereas low frequencies capture color and coarse structure. While white noise is, by design, information-free, Colorful-Noise introduces subtle, spatially varying color biases that guide the diffusion process toward outputs aligned with the structure and color of a reference image, while leaving the high-frequency components unconstrained, allowing the model to generate diverse fine details (see ~\cref{fig:teaser}).

\begin{figure}[t]
    \centering
    \includegraphics[width=\columnwidth]{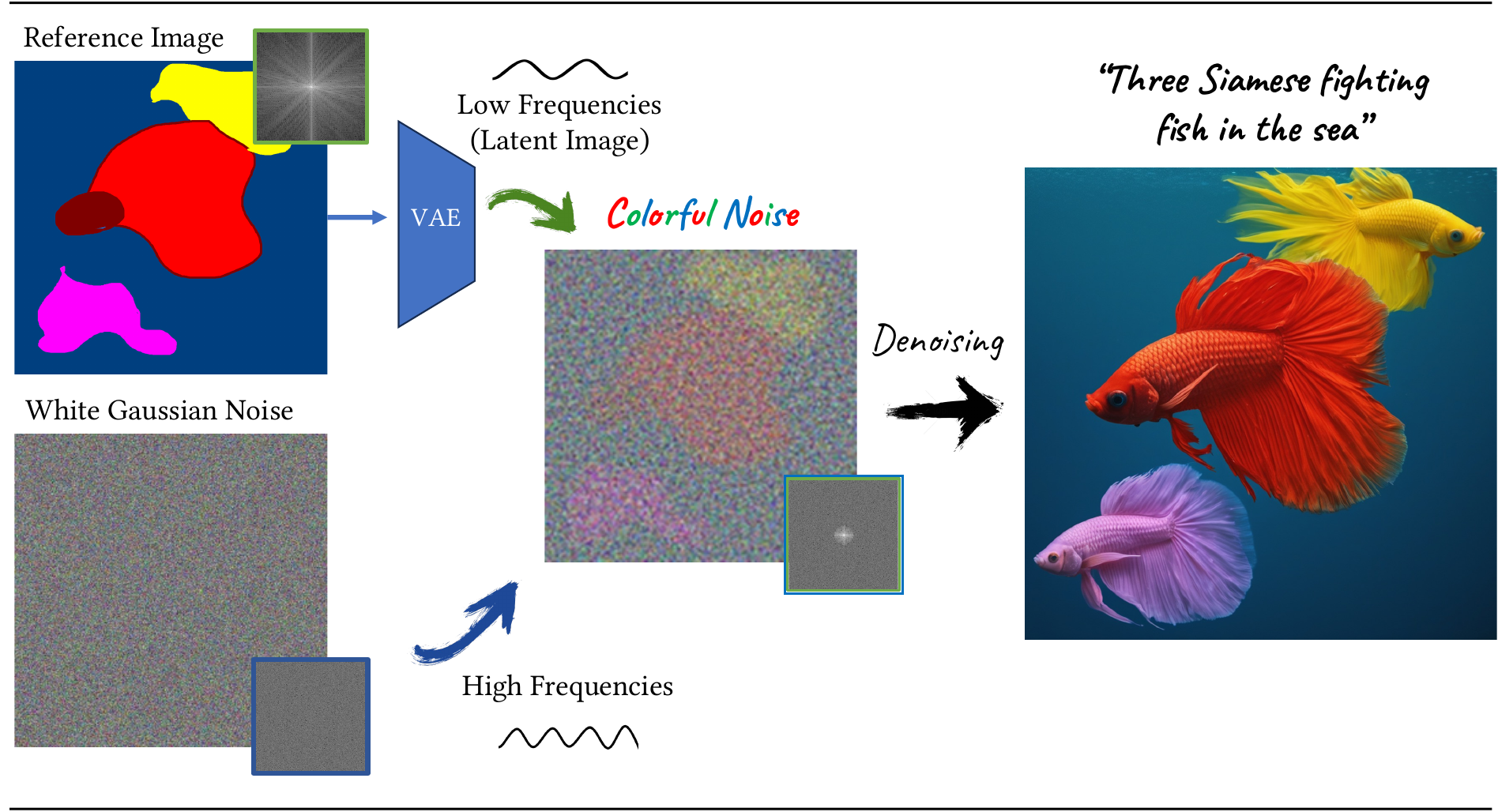}
    \caption{\textbf{Method Overview.} Colorful-Noise simply replaces the low frequency components of the \wgn\ with the low frequency of some conditioning reference image in latent space (in this case, a color stencil). Although the resulting noise is biased (i.e. not ``white''), it can still be used to successfully generate a high-quality image, while conditioning the desired results (note that the visualization here is in pixel space only for illustration).}
    \label{fig:method}
\end{figure}

Our experiments demonstrate that, when applied judiciously, low-frequency components of the noise latent can be directly manipulated to condition both the color and structure of the generated image, without requiring any additional training or optimization. Due to the low-level nature of this intervention, the conditioning signals can be easily handcrafted, yet still exert a strong influence on the output. This provides a lightweight alternative (see ~\cref{fig:method}) to existing image-based conditioning methods, which typically rely on algorithmically extracted signals from real images (e.g., Canny edges or depth maps) and are often sensitive to deviations from their expected input domain.

Furthermore, we observe that this frequency-based decomposition is not specific to a single architecture and generalizes across different latent spaces, including those of UNet- and flow-based generative models. Its focus on low-frequency components also allows it to be applied simultaneously with other high-frequency conditioning methods without interference.

We summarize our contributions as follows:
\begin{enumerate}
\item We analyze the frequency composition of \wgn\ latents and demonstrate a clear correspondence between latent frequency bands and semantic properties in the generated images, linking low frequencies to color and coarse structure, and higher frequencies to fine details.
\item We introduce Colorful-Noise, a simple, training-free method that conditions global structure and color using a lightweight and even handcrafted guidance image while preserving local details through biases injected into specific latent frequency bands.
\item We demonstrate how Colorful-Noise enables a wide range of creative applications, including flexible conditioning, color alignment, and stylistic image generation. 

\end{enumerate}

We share our code in our project page at: \link{\url{https://nadavc220.github.io/colorful-noise/}}.

\section{Related Work}

\begin{figure*}[t]
    \centering
    \includegraphics[width=\textwidth]{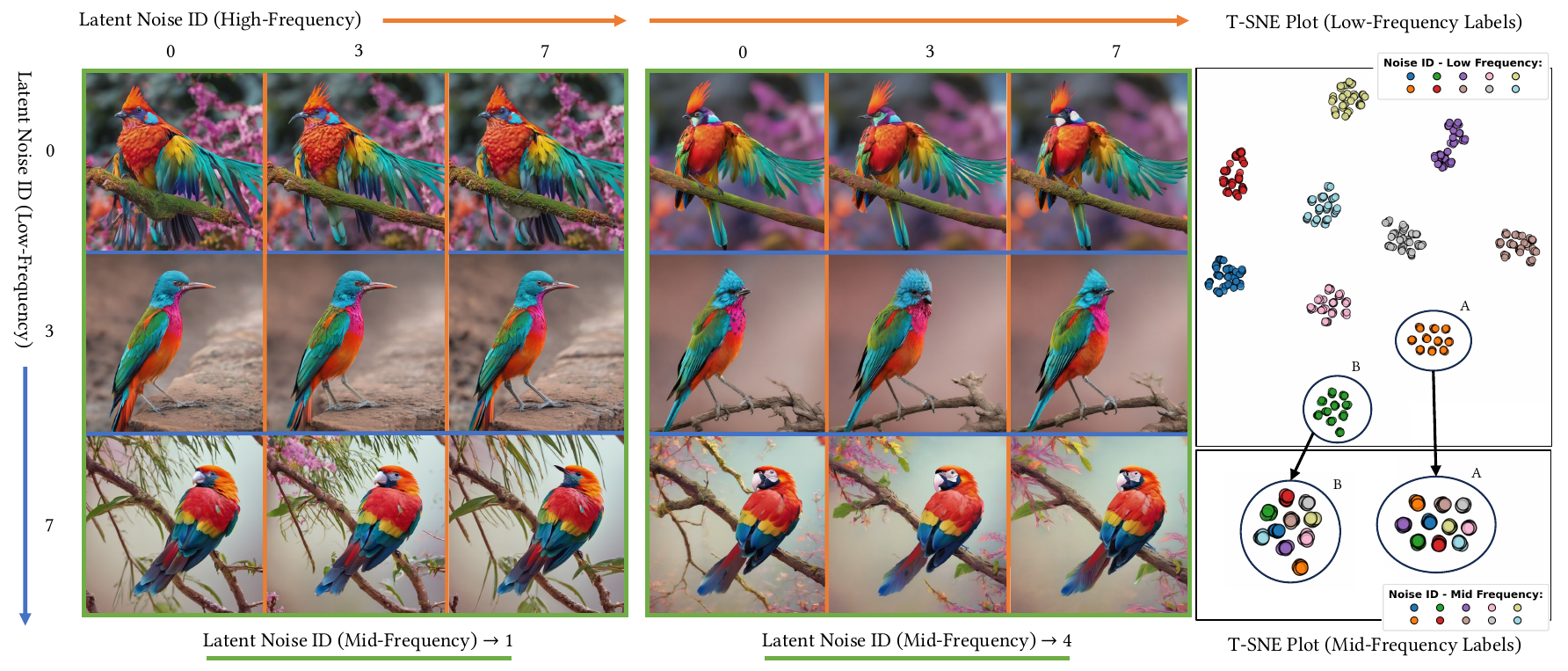}
    \caption{\textbf{Frequency Mixing in Noise Latents.} We decompose Gaussian noise latents into low, mid, and high frequency bands and recombine them to generate images from a fixed prompt (\textit{“A photo of a colorful bird”}). Rows share the same low frequencies, columns share the same high frequencies, and two mid-frequency variations are shown per pair. As can be observed, low frequencies dominate global structure and color, mid frequencies affect subtle details, while high frequencies have minimal visual impact, consistent with LPIPS-based t-SNE clustering as shown in the plot.}
    \label{fig:frequency_mix}
\end{figure*}

\paragraph{Text-to-Image Diffusion Models.}
Diffusion models are considered the state of the art for visual content generation. To produce an image—either in pixel space~\citep{saharia2022photorealistictexttoimagediffusionmodels, ramesh2021zeroshottexttoimagegeneration} or in a compressed latent space~\citep{rombach2021highresolution, podell2023sdxlimprovinglatentdiffusion, flux2024}—these models are trained to gradually denoise a random signal until a coherent image emerges. In text-to-image (T2I) models, this process is guided by a text prompt, allowing the recovery of fine details from isotropic Gaussian noise. Using normally distributed noise has become standard practice, giving the model maximal generative freedom~\citep{sohl2015deep}. Importantly, this noise provides no inherent structural priors—such as color, pose, or spatial layout—unless they are explicitly specified in the prompt. However, while the noise latent is central to diversity, it is typically treated as an opaque source of randomness rather than a controllable signal. Some prior work~\cite{blue_noise_in_diffusion, inverse_heat, nonisotropic} experiment with diffusion using non-white noise, like blue-noise~\cite{blue_noise} which has no energy in its low-frequencies. We further discuss and experiment with blue-noise in the supplemental file.


\paragraph{Conditional Image Generation.}

Following the success of Classifier-Free Guidance (CFG)~\citep{Ho2022ClassifierFreeDG}, many works introduce additional conditioning signals to control image generation. Most approaches inject structure through explicit feature conditioning, either by training auxiliary networks or modifying intermediate representations.

Methods such as ControlNet~\citep{zhang2023adding}, IP-Adapter~\citep{ye2023ip-adapter}, and GLIGEN~\cite{li2023gligen} leverage learned adapters to incorporate spatial control signals, while~\citet{avrahami2023spatext} enables region-specific text conditioning via segmentation masks. For more abstract guidance, style-based methods inject reference image features through attention manipulation~\citep{Hertz_2024_CVPR,wang2024instantstyle}, LoRA fine-tuning~\cite{frenkel2024implicitstylecontentseparationusing, shah2023ZipLoRA}, or iterative optimization~\cite{jiang2025balanced}.

While effective, these approaches typically require additional training, task-specific design, or careful balancing between competing modalities. When multiple conditioning signals are combined, misalignment and interference often arise due to differences in abstraction level and spatial specificity~\cite{Cohen_2025_CVPR}. In contrast, our approach introduces conditioning at the noise level, prior to denoising, avoiding competition between modalities.

\paragraph{Noise-Manipulation in Image Generation.}
Implicit noise manipulation leverages the latent noise variables indirectly, typically through optimization or inversion. Diffusion inversion methods~\citep{garibi2024renoise,miyake2025negative,tumanyan2023plug,mokady2023null,huberman2024edit,zhang2023real} aim to recover a noise latent that reconstructs a given image, enabling subsequent editing using the pretrained generative prior. While powerful for reconstruction, these methods are often have limited editing abilities.

Explicit noise manipulation directly modifies the noise distribution before or during denoising. Examples include spatially varying denoising schedules~\citep{levin2025differential} or palette-based latent mixing for global color alignment~\citep{Shum_2025_CVPR}. Although effective within their scope, these methods are typically task-specific and do not generalize across different forms of conditioning.

In contrast, we show that direct, frequency-selective manipulation of the initial noise latent provides a simple and general mechanism for conditioning color and structure, without optimization, auxiliary networks, or task-specific training.




\paragraph{Frequency-Aware Generation Methods.}
Several recent works explore frequency-aware mechanisms in diffusion models~\citep{zhang2024frecas,liu2025one,ren2025fds,hertz2023delta}. These methods typically leverage frequency information during intermediate denoising steps to improve image quality or enable localized editing. For example, FDS~\citep{ren2025fds} performs optimization-based manipulation of specific frequency bands using a DDS objective~\citep{hertz2023delta}. While effective, it requires manual band selection and introduces significant inference overhead.

In contrast, our method operates directly on the initial noise latent, requires no optimization, and relies on a simple and interpretable frequency decomposition. This enables lightweight, composable conditioning that generalizes across architectures and can be combined seamlessly with other control mechanisms.






\section{White Gaussian Noise Analysis}
\label{sec:analysis}

We analyze how the frequency structure of the initial latent \wgn\ affects SDXL image generation. Unlike prior work that tracks attribute evolution over diffusion timesteps~\cite{beta_is_all, issachar2025dypedynamicpositionextrapolation}, we focus on correlations between visual attributes and the Fourier decomposition of the input noise. Following the SDXL setup, we use four-channel \wgn\ latents of size $128\times128$ transformed to a frequency representation using the Fast Fourier Transform~\cite{fft}.

We begin by defining operators for frequency extraction, and then analyze two main perspectives: (1) how noise frequencies at different levels correlate with the visual properties of the generated outputs, and (2) how manipulating these frequencies using an external signal affects those visual properties. 

\subsection{Fourier Frequency Extraction}
Given a latent $z \in \mathbb{R}^{C \times H \times W}$, we define $\hat{z} \in \mathbb{C}^{C \times H \times W}$ as the frequency representation of z, calculated separately for each channel, using the 2D Discrete Fourier transform (DFT) and $\hat{z}_L, \hat{z}_M, \hat{z}_H$ as the corresponding signals with only the low-, mid-, and high frequencies of $\hat{z}$, respectively. For convenience, we define $\mathcal{F}$ as the combined operation of DFT and frequency extraction, and $\mathcal{F}^{-1}$ as the inverse operation of re-combining the frequencies and applying the inverse Discrete Fourier Transform as follows:

\begin{equation}
\begin{aligned}
\hat{z}_L, \hat{z}_M, \hat{z}_H = \mathcal{F}(z; \alpha, \beta), \\
z = \mathcal{F}^{-1}(\hat{z}_L, \hat{z}_M, \hat{z}_H),
\end{aligned}
\end{equation}

where $\alpha,\beta\in[0,1]$ define the low ($<\alpha$) and high ($>\beta$) cutoffs.

\subsection{Latent Frequency Bands Influence on Generation}
\label{sec:analysis_birds}
We sample $n=10$ latents $\{z^i\}_{i=1}^n$, and decompose each using the values of $\alpha=0.25,\beta=0.75$,

\begin{equation}
[\hat{z}_L^i, \hat{z}_M^i, \hat{z}_H^i] = \mathcal{F}(z^i; \alpha, \beta),
\end{equation}

and form all $n^3$ mixed latents:

\begin{equation}
\label{eq:idft_experiment}
z_{(i,j,k)}=\mathcal{F}^{-1}(\hat{z}_L^i, \hat{z}_M^j, \hat{z}_H^k),\;\; 1\le i,j,k\le n.
\end{equation}

We generate $n^3$ images from these latent noise samples using a fixed prompt, and then cluster them using pairwise perceptual distances, which are measured using LPIPS~\cite{lpips}. As shown in \cref{fig:frequency_mix}, results cluster primarily by the low-frequency index, while high-frequency choices are dispersed. Mid-frequencies mainly induce within-cluster variation, affecting fine details. Overall, low frequencies dominate global structure and color distribution, whereas mid frequencies modulate finer appearance. We repeat this experiment for additional prompts and report quantitative results in the supplemental file.

\subsection{Compositing Natural Frequency into Noise}

Given an encoded natural image $I$ and a noise latent $z$, we independently inject frequency components from $I$ into $z$ to isolate their individual effects. Specifically, as we replace the low-frequency band of the noise latent with the corresponding frequency band from the image, while keeping the remaining bands unchanged:

\begin{equation}
\begin{aligned}
z^I_L &=  \mathcal{F}^{-1}(\gamma \hat{I}_L, \hat{z}_M, \hat{z}_H), \\
\end{aligned}
\end{equation}

Where $\gamma\in\mathbb{R}$ is a scaling-factor introduced to mitigate the distributional shift between the natural image frequencies and those of the white Gaussian noise input.   


We use the Aesthetic-4K~\cite{zhang2025diffusion4k} dataset as a source of natural images and replace lowest 25\% frequency band to prevent leakage between groups. To prevent information from being conveyed through text conditioning, we use an empty prompt. Sample results of this experiment are shown in \cref{fig:analysis_injection}. As observed,  replacing the low-frequency components of the noise—when properly $\gamma$ scaling-enables the model to generate images that inherit the low-frequency structure of the conditioning image suggesting that low frequency can condition the generated image low-frequencies without shifting out of the optimized noise distribution.

\begin{figure}[t]
    \centering
    \includegraphics[width=\columnwidth]{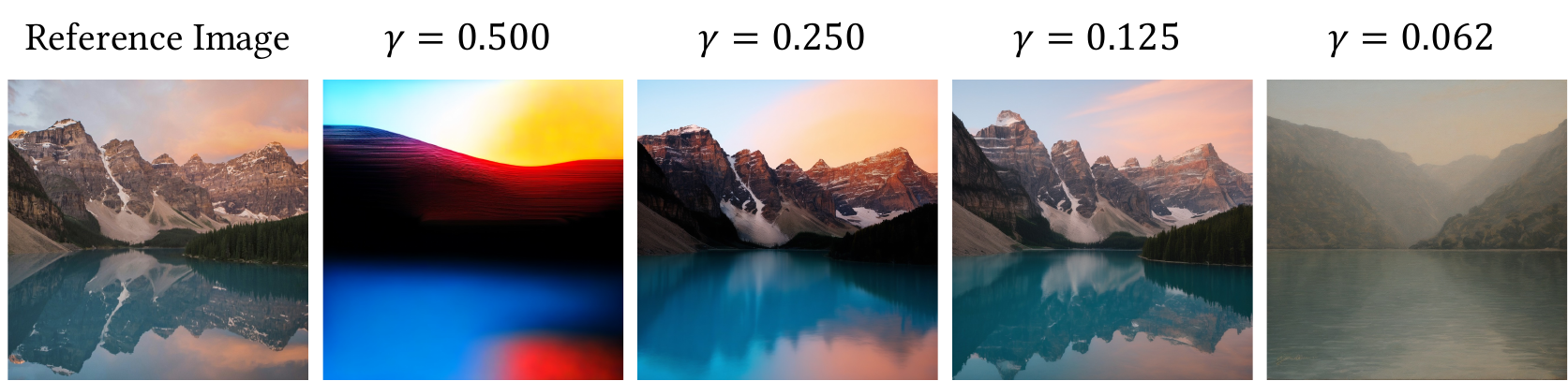}
    \caption{\textbf{Noise and Image Combinations.} We generate images by mixing noise with the low-frequency components of natural images. As observed, replacing low frequencies allows for a surprising reconstruction of the reference image’s low-frequency content—even without a prompt—when appropriately scaled. (Zoom in for a better view.)}
    \label{fig:analysis_injection}
\end{figure}

We also experiment with replacing the mid- and high-frequencies components, and show they fail to condition the image. We share more details and quantitative evaluations in the supplemental file.


\begin{figure*}
    \centering
    \includegraphics[width=\textwidth]{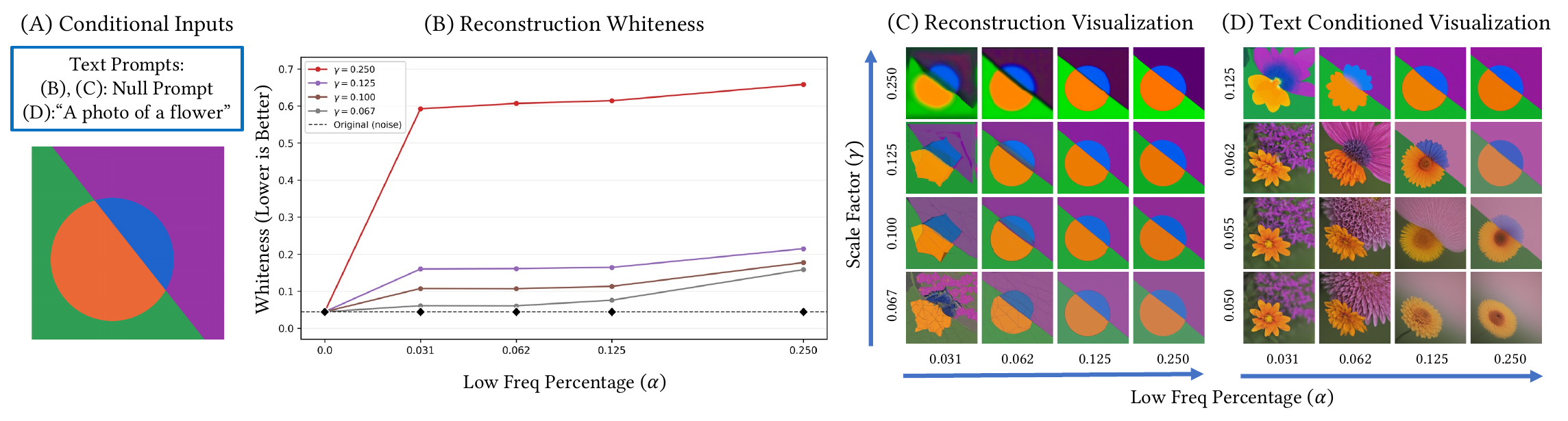}
    \caption{\textbf{$\alpha$ and $\gamma$ Ablation Study.} We ablate the low-frequency ratio $\alpha$ and scale $\gamma$ using synthetic low-frequency inputs. (A) Conditional inputs. (B) PSD-based whiteness shows that smaller $\alpha$ and $\gamma$ better preserve spectral balance. (C) Promptless generation indicates that large $\alpha$ or $\gamma$ introduce color and structural artifacts, while smaller values preserve low-frequency color. (D) With text conditioning, high $\alpha$ overconstrains structure and high $\gamma$ causes blur, whereas smaller $\gamma$ yields a better balance between color, structure, and prompt alignment.}
    \label{fig:frequency_rec}
\end{figure*}

\section{Method - Colorful-Noise}
\label{sec:method}

Following our findings in \cref{sec:analysis}, we formulate a simple method to control low-frequency visual characteristics of generated images in diffusion by manipulating the corresponding frequencies in the latent noise. We apply Fourier transform to both a noisy latent and input conditional image, then replace the low-frequency components of the noisy latent with those of the conditional image before applying the inverse Fourier transform.

Let $C$ be a conditioning RGB image, and let $z \sim \mathcal{N}(\mathbf{0}, \mathbf{1})$. Since the denoising process of $z$ occurs in the VAE latent space, we first encode $C$ into this space:

\begin{equation}
c = \mathcal{E}(C),
\end{equation}

where $\mathcal{E}$ represents the VAE encoder. Next, to condition $z$ on $c$, we decompose both latents using the function $\mathcal{F}$:

\begin{equation}
\begin{aligned}
\hat{c_L}, \hat{c_M}, \hat{c_H} &= \mathcal{F}(c; \alpha, \beta) \\
\hat{z_L}, \hat{z_M}, \hat{z_H} &= \mathcal{F}(z; \alpha, \beta)
\end{aligned}
\end{equation}
where $\beta=1$ as we do not separate mid and high frequencies.
Conditioning is applied by combining the low-frequency components of latent conditioning $c$ with the mid- and high-frequency components of latent noise $z$ :

\begin{equation}
z^c = \mathcal{F}^{-1}(\gamma \hat{c}_L, \hat{z}_M, \hat{z}_H).
\end{equation}

$\gamma$ is the scaling factor that controls the strength of the conditioning image in latent noise. 
When appropriately scaled, the modified latent appears nearly identical to the human eye but encodes a subtle bias towards the structure and color of the conditional image, a delicate alteration that nonetheless has a strong influence on the final output. We refer to this technique as \emph{Colorful-Noise} (see ~\cref{fig:method}) akin to colored-noise (e.g.\ pink/brown/blue noise), which is noise that has a specific frequency distribution. 



As an alternative, we also experiment with Discrete Wavelet Decomposition, which has gained popularity for image frequency decomposition. We find that it performs comparably to the Fourier-based approach, though it is less flexible for decomposing the image into finer sub-bands. Details and examples are provided in the supplemental material.

\subsection{Choosing $\alpha$ and $\gamma$}

As shown in \cref{sec:analysis}, scaling the energy of injected frequencies alleviates the whiteness shift caused by the statistical mismatch between natural images and \wgn. To further analyze the roles of the low-frequency injection ratio $\alpha$ and scaling factor $\gamma$, we perform a focused ablation study.

Since our analysis targets low-frequency injection, we construct a synthetic low-frequency dataset of 1K flat images with random colors and simple shapes. This controlled setup isolates the effect of low-frequency color and structure (see supplemental file). We quantify whiteness shift using Power Spectral Density (PSD), defining whiteness as the standard deviation of band-wise power across channels, where lower values indicate more uniform spectra.

We evaluate colorful-noise latents under two settings: (i) an empty prompt to assess low-frequency reconstruction, and (ii) a text prompt to evaluate editability and semantic alignment. We sweep $\alpha$ and $\gamma$, with results shown in \cref{fig:frequency_rec}.

For promptless generation (B, C), smaller $\alpha$ and $\gamma$ better preserve spectral whiteness, while larger $\alpha$ values introduce imbalance and visible color and structural artifacts. Increasing $\gamma$ further amplifies these effects, leading to blur and reduced color fidelity. Combining low $\alpha$ and low $\gamma$ preserves color while relaxing shape constraints, resulting in mild geometric distortions that favor conditioning over exact reconstruction.

Similar trends are observed for text-conditioned generation (D): large $\alpha$ overly constrains global structure, while large $\gamma$ degrades color and sharpness. In contrast, smaller $\gamma$ values provide an effective balance between low-frequency conditioning and semantic alignment.







\section{Applications}
\label{sec:applications}
Based on our analysis in \cref{sec:analysis}, we find that when used alongside a text prompt, Colorful-Noise guides the output image’s colors and general structure according to a colorful reference image, while the text prompt determines the semantic content.

\begin{figure*}[t]
    \centering
    \includegraphics[width=\textwidth]{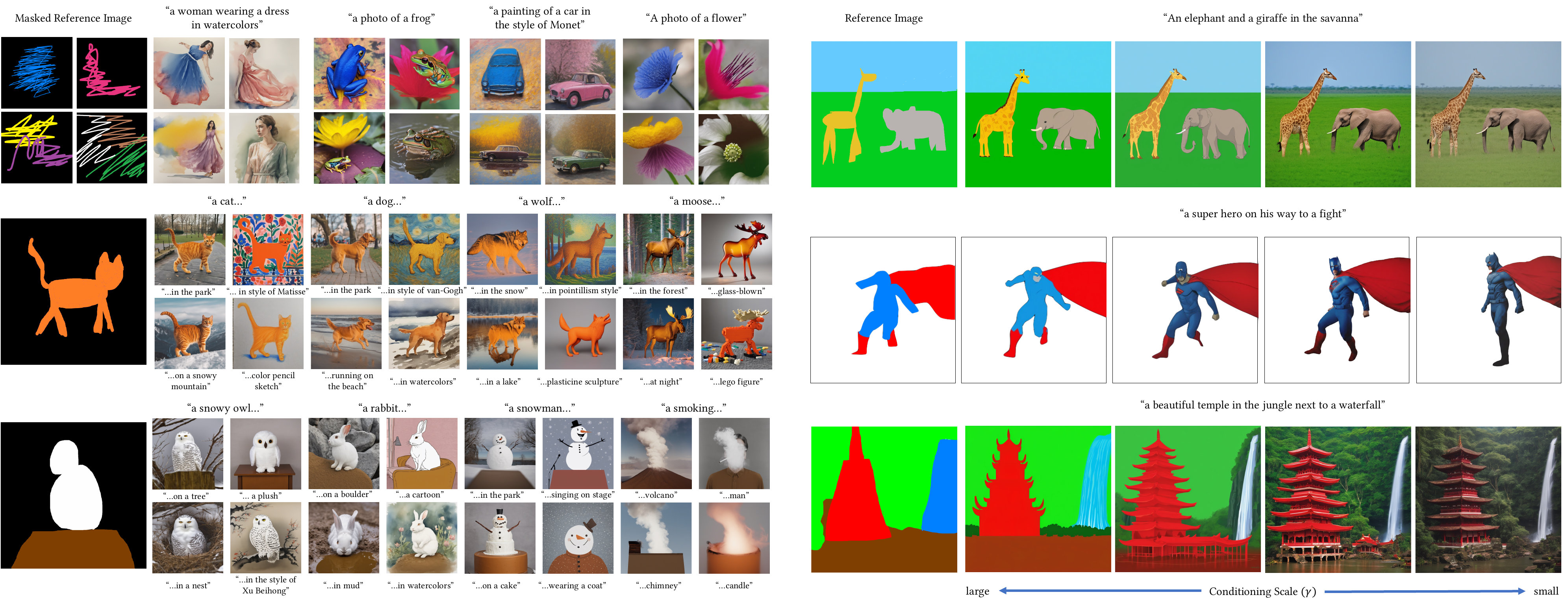}
    \caption{\textbf{Localizing Colorful-Sketch Conditioning.} On the left we demonstrate how a single masked condition can generate color/structure aligned results for various prompts.
    The top row shows four simple masked scribble conditions (the black background is masked out)  and the four corresponding results for various prompts.
    The middle and bottom rows show two simple masked sketch conditioning various prompts. 
    On the right we show the effect of changing the scale factor for full-conditioned examples (without masking). Zooming in for a better view or refer to the supplemental file for larger figures. 
    }
    \label{fig:localizing}
\end{figure*}

\begin{figure*}
    \centering
    \includegraphics[width=\textwidth]{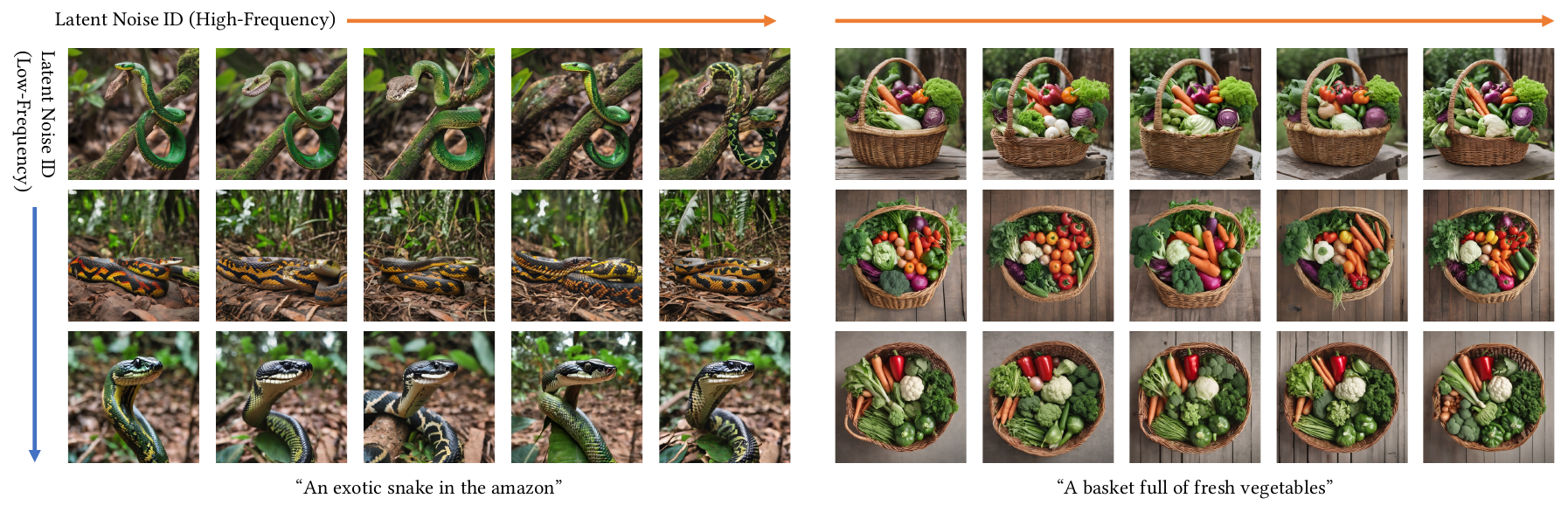}
    \caption{\textbf{Hierarchical Exploration} of image generation by separating low- and high-frequency noise. (Zoom in to observe variations in fine-details.) }
    \label{fig:exploration}
\end{figure*}

In this section, we explore creative applications of this approach using color maps at increasing levels of detail—from simple scribbles, through coarse semantic sketches, to real photographs—while combining them with additional conditional inputs. Although explicitly specifying colors in the text prompt can enhance the semantic mapping of colors to subjects (see \cref{subsec: ablation}), we omit such references in our examples to highlight the effectiveness of Colorful-Noise alone. Our experiments focus on SDXL, but to assess the generality of Colorful-Noise, we also present various results on the flow-based model Flux‑dev1.0~\cite{flux2024} in \cref{fig:flux_results}. The results were obtained using an average of two parameter-tuning iterations for SDXL, and an average of six parameter-tuning iterations on Flux-dev1.0. 

\subsection{Text-Only Generation-Space Exploration}
The simplest application of our method allows the separation of low and high fequencies to explore alternative image generations conditioned only on a text prompt. 
As shown in \cref{fig:exploration}, fixing the low frequencies of the noise while randomizing the high frequencies allows one to create variation in fine details, while preserving the overall color scheme and structure of the image. This enables heirarchical exploration of the generation space: users can first select a global appearance of structure and colors by exploring different alternative low-frequency noise, and then refine it by varying only the high-frequency details. 

\subsection{Sketch Conditioned Generation}
We condition \wgn\ using hand-drawn colorful sketches that convey varying semantic detail, with example results shown in \cref{fig:localizing,fig:flux_results}. We consider two settings: (1) Masked Conditioning, where colorful noise replaces the original noise only in painted regions, and (2) Full Conditioning, where the full colorful-noise latent is used. Since the conditional images are dominated by low frequencies due to their limited detail, we set $\alpha=0.03$ and $\gamma \approx 0.04$, replacing only a small portion of low-frequency noise while applying strong normalization to preserve latent noise whiteness.

Masked conditional latents (\cref{fig:localizing}, left) produce plausible images despite combining colorful and white noise, for both simple color scribbles and more semantic sketches. Moreover, a single conditional can adapt to multiple text prompts, with the model effectively balancing color guidance and textual semantics so that both influences are clearly reflected in the result. The same fidelity is observed under full conditioning (\cref{fig:localizing}, right): higher conditioning scales emphasize color guidance, producing stylized cartoon-like results, while smaller $\gamma$ values yield more photorealistic images.

\subsection{Image Conditioned Generation}
While colorful sketches provide a convenient form of conditioning, we find that using natural images to condition the noise also enables compelling applications. For this setting, we use $\alpha=0.125$ and $\gamma \approx 0.2$, since natural images are less dominated by low frequencies than colorful sketches, and require less scaling to preserve whiteness. 

\paragraph{\textbf{Color-Based Style Alignment.}} We apply colorful noise conditioned on natural images across multiple prompts. As in the colored-sketch case, a single conditional yields diverse results by adaptively balancing text semantics and color alignment. This provides an effective way to generate series of color-aligned images. Results are shown in \cref{fig:alignment,fig:flux_results}, and in the supplemental file.

\begin{figure*}
    \centering
    \includegraphics[width=\textwidth]{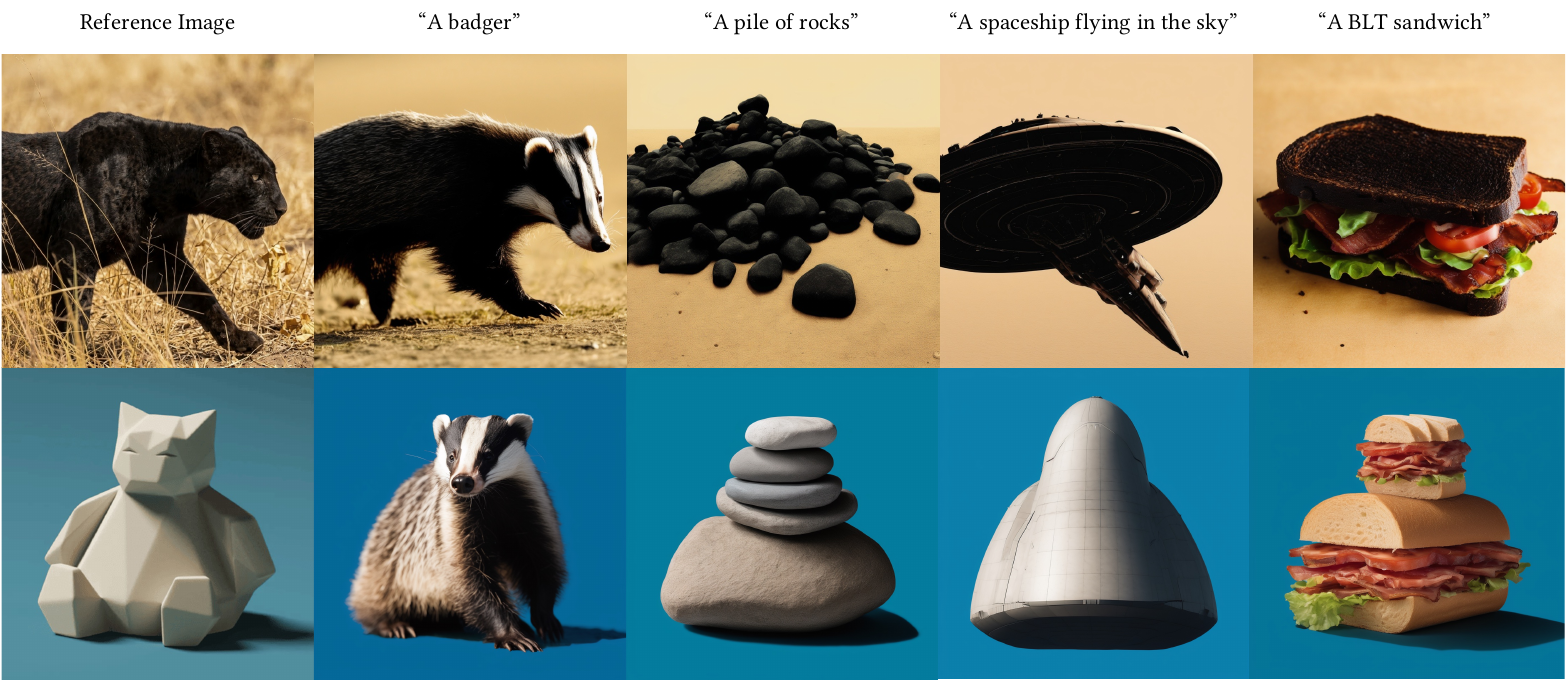}
    \caption{\textbf{Color-Based Style Alignment.} Results for Color-Based style alignment using SDXL. Bottom reference © Augustin Arroyo (@flowalistic on Instagram). All rights reserved.}
    \label{fig:alignment}
\end{figure*}

\begin{figure*}
    \centering
    \includegraphics[width=\textwidth]{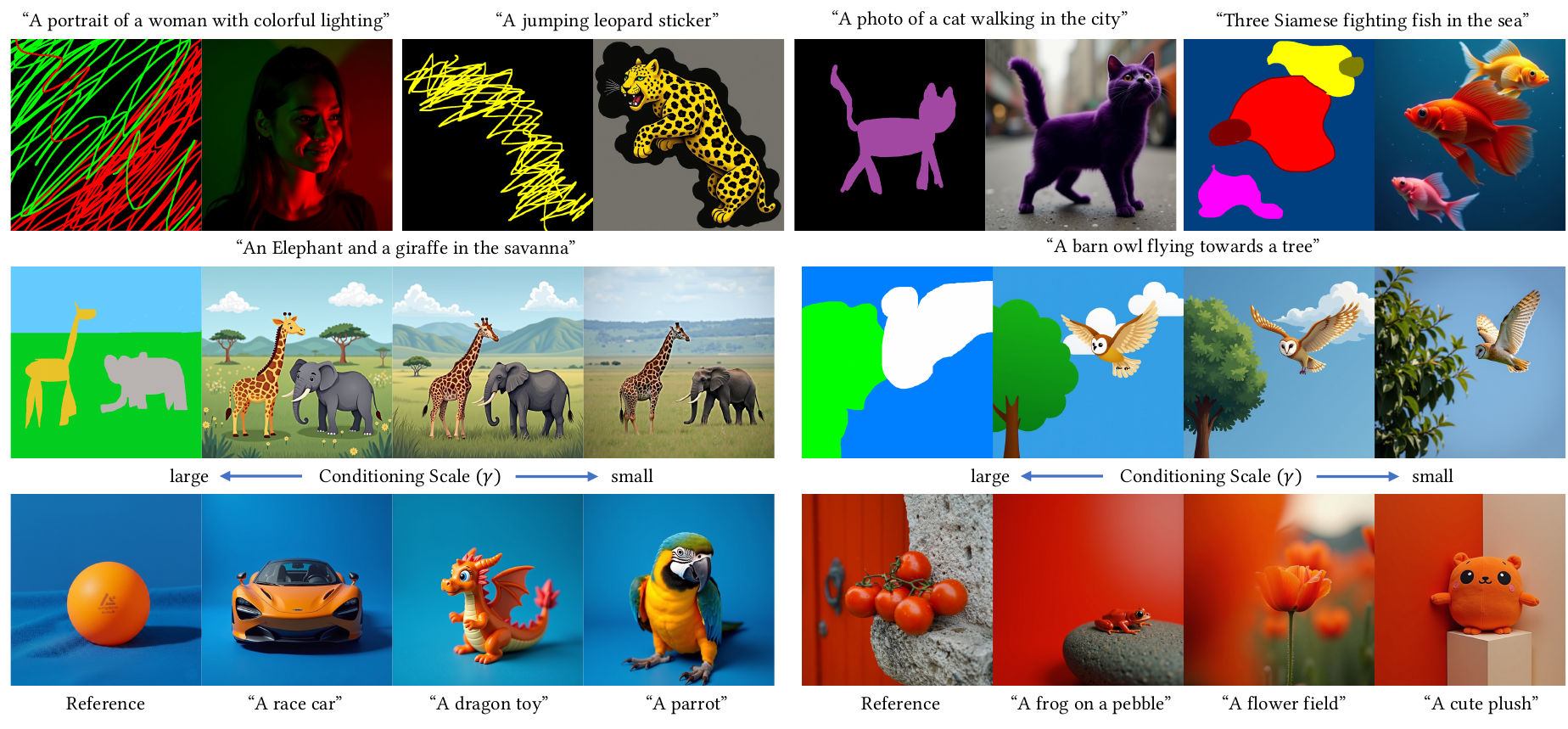}
    \caption{\textbf{Colorful-Noise with Flux:} results of various applications using Flux Flow-based model.}
    \label{fig:flux_results}
\end{figure*}

\begin{figure*}
    \centering
    \includegraphics[width=\textwidth]{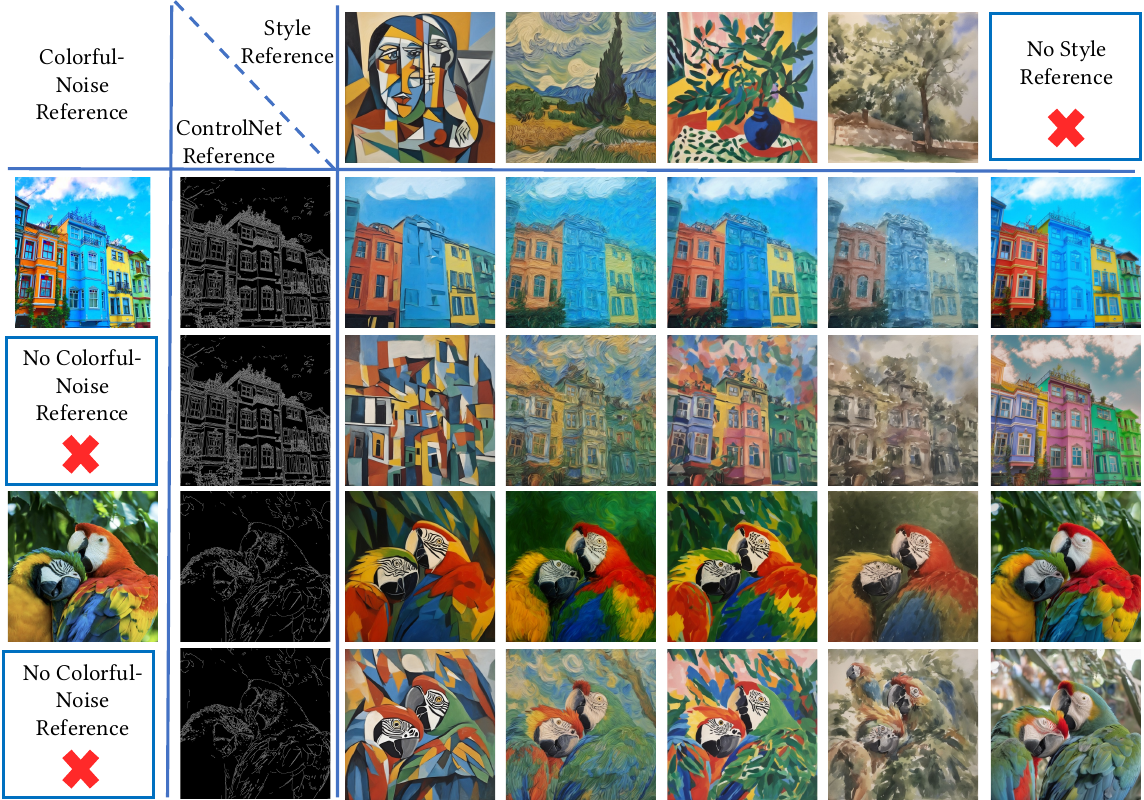}
    \caption{{\textbf{Color-Preserving Stylization.} Combining low-frequency conditioning of Colorful-Noise with high-frequency conditioning of structure and style conditioning. See text for detailed explanation. Prompts: \textit{"Colorful buildings"} (Top), \textit{"Two Macaws"} (Bottom).}}
    \label{fig:stylization}
\end{figure*}

\begin{figure*}
    \centering
    \includegraphics[width=\textwidth]{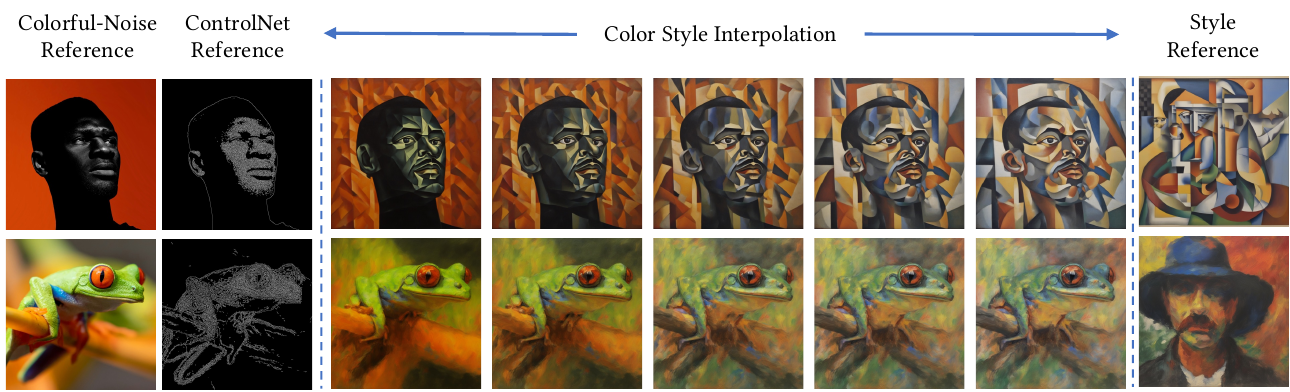}
    \caption{{\textbf{Color-Style Interpolation.} Color-Style Interpolation. Colorful noise can be partially applied via linear interpolation with white noise. Combined with style and Canny conditioning, this produces a smooth transition between the colors of a reference image (right) and a style image (left), while preserving other aspects of the style reference. Prompts: \textit{"A portrait of a man"} (Top), \textit{"A frog on a branch"} (Bottom).}}
    \label{fig:stylization_interp}
\end{figure*}

\paragraph{\textbf{Color-Preserving Stylization.}} By design, Colorful-Noise conditions the output using low-frequency information from a reference image. This allows combining it with methods for high-frequency control. We show results of using Colorful-Noise alongside ControlNet~\cite{zhang2023adding} to guide fine details with a Canny map, and Conditional-Balanced StyleAligned~\cite{Hertz_2024_CVPR, Cohen_2025_CVPR} for image reconstruction and stylization (see \cref{fig:stylization,fig:stylization_interp} and supplemental file). 

When combined with ControlNet, Colorful-Noise can generate photographs closely matching the conditional image (No Style Reference column). Adding style conditioning demonstrates that Colorful-Noise preserves color without compromising other stylistic aspects, such as geometry or strokes, and can even enhance structural understanding, improving high-frequency generation. In some cases, semantic information is lost without color conditioning but retained with it. Additionally, linearly interpolating between colorful and white noise allows smooth interpolation between the color styles of the style and colorful-noise reference images.

\section{Additional Experiments}
To further examine the capabilities of Colorful-Noise, we conduct additional experiments, comparing its conditioning effectiveness with existing methods and performing an ablation study to analyze how varying levels of text prompt influence the output when combined with Colorful-Noise.

\subsection{Comparisons with Prior Works}

For quantitative evaluation, we consider three color-preservation tasks: (1) \textit{Image Variation}, where a raw image is used to generate a structurally and chromatically similar output; (2) Colorfield-to-Image, where low-resolution color patches guide local colors; and (3) \textit{Image-to-Image Color Transfer}, where only the global color distribution is preserved. In each setting, we compare against the most relevant prior methods for that task.

We report CLIPScore~\citep{Hessel2021CLIPScoreAR} and two variants of Earth Mover’s Distance (EMD)~\citep{710701}, which measures distance between distributions. To evaluate structural preservation and local color consistency, we compute EMD over corresponding image patches (task-dependent), referred to as \textit{Localized-EMD}. We additionally report \textit{Global-EMD} as a reference, although our method does not explicitly target global color appearance.

Experiments are conducted on 195 images from the Aesthetic-4K [Zhang et al. 2025] evaluation set with captions. All results in Tab. 1 use a resolution of 512 × 512 and a patch size of 64 × 64. Additional visual comparisons are provided in the supplementary material.

For Image Variation, vanilla SDXL performs poorly since color cues rely solely on text. Methods such as~\citet{Shum_2025_CVPR}-ZS and T2I-Adapter~\citep{10.1609/aaai.v38i5.28226} improve color preservation but degrade image–text alignment. In contrast, our method maintains alignment while achieving significantly better L-EMD scores.

In the Colorfield-to-Image task, our approach improves local color–structure consistency compared to~\citet{Shum_2025_CVPR}-ZS, while remaining competitive in global color preservation. Although \citet{BRIA=ControlNet} achieves stronger L-EMD, it suffers from reduced text alignment and tends to overfit to color grids, producing visible artifacts (see supplemental Figure 3).

Finally, in Image-to-Image Color Transfer, our method—despite not being explicitly designed for this task—achieves better local color consistency than~\citet{Shum_2025_CVPR}-ZS and StyleAligned~\cite{Hertz_2024_CVPR}, demonstrating robustness across diverse inputs.

Overall, performance decreases as guidance becomes less restrictive: from full structure and color (Image Variation), to local color (Colorfield-to-Image), and finally to global color distribution (Color Transfer).

\renewcommand{\arraystretch}{1.1} 

\begin{table}[ht]
\footnotesize
\centering
\caption{\textbf{Comparisons with Prior Works.} For each task, we denote results in \textbf{bold} and \underline{underline} as the best and second-best performance, respectively. Among all methods, only Colorful-Noise incur no additional memory or runtime overhead over the SDXL baseline.
}
\label{tab:comparisons}
\small
\resizebox{\columnwidth}{!}{%
\begin{tabular}{l|cccc}
\hline
\multirow{2}{*}{Method}       & \multirow{2}{*}{\begin{tabular}[c]{@{}c@{}}Guidance\\ Type\end{tabular}} & \multirow{2}{*}{CLIPScore $\uparrow$} & \multicolumn{2}{c}{EMD $\downarrow$} \\ \cline{4-5} 
                              &                                                                          &                                       & Localized         & Global           \\ \hline
\textit{Color Variation}      &                                                                          &                                       &                   &                  \\
\hspace{3mm}SDXL              & Text only                                                                & \textbf{34.84}                        & 20.08             & 12.91            \\
\hspace{3mm}T2I Adapter-Color & Inversion                                                                & 33.68                                 & {\ul 14.14}       & 10.46            \\
\hspace{2mm} \protect~\citet{Shum_2025_CVPR}-ZS        & Raw Image                                                                & 32.25                                 & 17.40             & \textbf{2.10}    \\
\hspace{3mm}Ours              & Low Freq.                                                                & {\ul 33.68}                           & \textbf{8.84}     & {\ul 5.47}       \\ \hline
\textit{Colorfield to Image}  &                                                                          &                                       &                   &                  \\
\hspace{3mm}ControlNet-Color  & Colorfield                                                               & 27.95                           & \textbf{7.30}     & {\ul 3.64}       \\
\hspace{2mm} \protect~\citet{Shum_2025_CVPR}-ZS        & Colorfield                                                               & {\ul 29.31}                                 & 15.83             & \textbf{2.27}    \\
\hspace{3mm}Ours              & Low Freq.                                                                & \textbf{32.68}                        & {\ul 10.12}       & 6.74             \\ \hline
\textit{I2I Color Tarnsfer}   &                                                                          &                                       &                   &                  \\
\hspace{2mm} \protect~\citet{Shum_2025_CVPR}-ZS        & Raw Image                                                                & 30.68                                 & 25.46             & 18.18            \\
\hspace{3mm}StyleAligned      & Raw Image                                                                & {\ul 31.87}                           & {\ul 22.47}       & {\ul 16.21}      \\
\hspace{3mm}Ours              & Low Freq.                                                                & \textbf{34.90}                        & \textbf{19.91}    & \textbf{12.55}   \\ \hline
\end{tabular}
}
\end{table}

\subsection{Ablation Study}
\label{subsec: ablation}

\begin{figure}[t]
    \centering
    \includegraphics[width=\columnwidth]{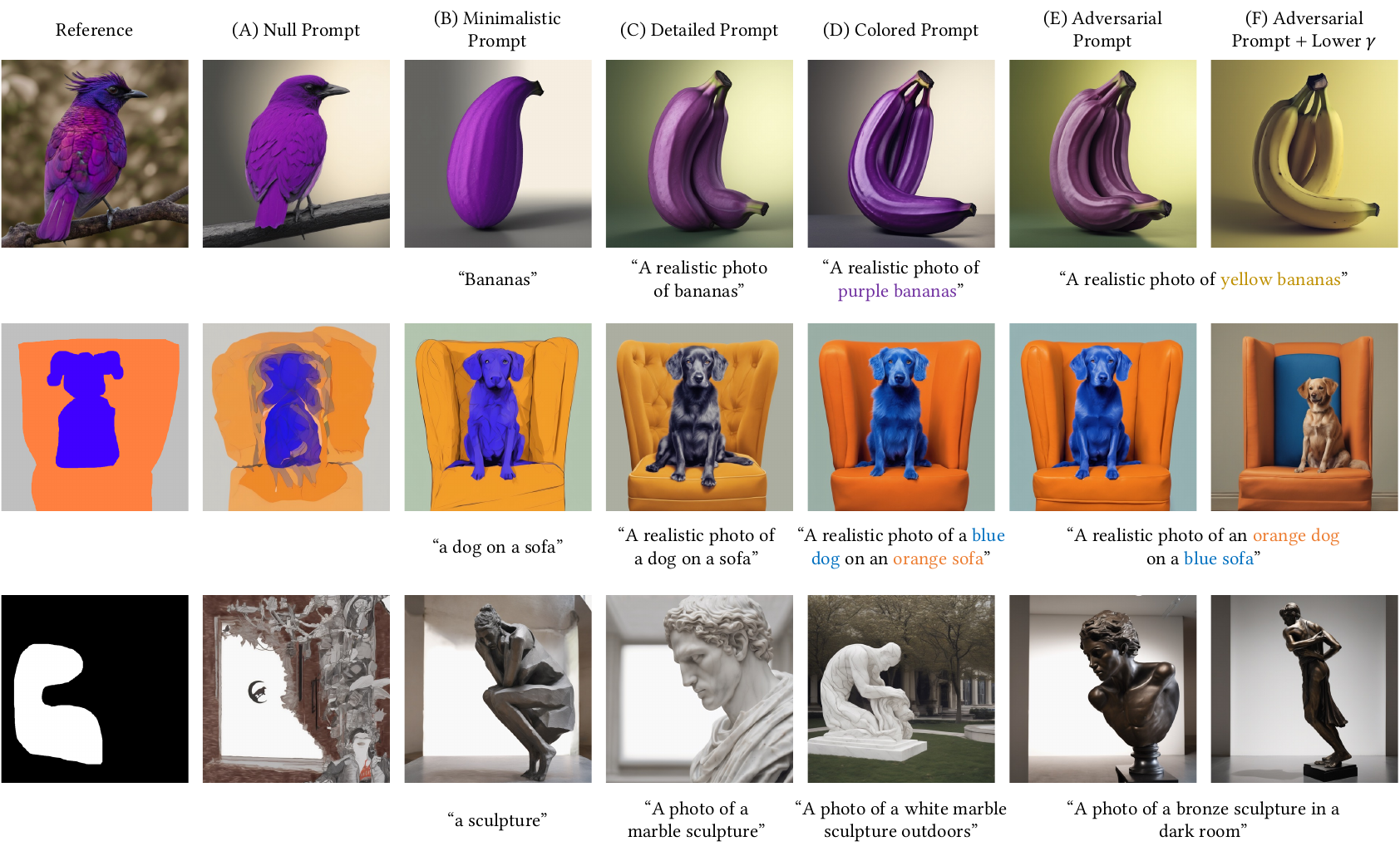}
    \caption{\textbf{Prompt Ablation Study.} We investigate the effect of various prompt levels on the output, when conditioned with a photo, a full colorful-sketch, and a masked colorful-sketch. As there is no semantic matching between colors and subject in the prompt, the generation model is free to interpret the color mapping freely which leads to various semantic interpretations based on the input prompt. See higher resolution version in supplemental file, Figure 7.}
    \label{fig:frequency_edit}
\end{figure}

To analyze the effect of prompt variation, we evaluate three image conditions—photograph, full sketch, and masked sketch—paired with five prompt types: (A) null, (B) minimal, (C) detailed, (D) color-aware, which specifies color–subject relationships from the guidance image, and (E) adversarial, assigning conflicting colors. For the adversarial case, we generate an additional example with a lower conditioning scale $\gamma$ (F). All images are generated using the $\alpha$ and $\gamma$ values from \cref{sec:applications}.

Null and minimal prompts often produce implausible results, whereas detailed prompts (C) yield more interesting behavior, particularly for sketches. Without photographic constraints, the model may reassign colors (e.g., mapping the subject to dark blue and using the mask for the background), revealing two limitations of Colorful-Noise: its flexibility can lead to deviations from the intended condition in out-of-distribution cases, and in some cases, colors may be interpreted contrary to user intent, especially with masked inputs. In contrast, color-aware prompts (D) enforce both semantic consistency and mask adherence.

In adversarial settings, unmasked conditioning strongly enforces color, while masked conditioning allows the colored region to take on a different semantic role aligned with the prompt (E). Lowering the conditioning scale improves prompt alignment even in adversarial cases, demonstrating that Colorful-Noise can preserve structural layout independently of strict color enforcement (F).
While this effect appears in some cases, lowering $\gamma$ can also degrade both structure and color. We hypothesize this depends on how strongly structure is encoded in the frequency domain—when weak, reducing $\gamma$ harms both simultaneously.


\section{Discussion}
In this work, we present Colorful-Noise, a simple yet effective method for biasing \wgn\ latents toward specific colors and structural cues that persist throughout the diffusion process. Our approach is motivated by analyzing the relationship between frequency components in Gaussian noise and the resulting generated images, revealing a direct connection despite differences in spectral power density. We show that even subtle manipulation of low-frequency components enables creative applications such as hierarchical image exploration and structural and stylistic conditioning. Unlike prior training-based methods, our approach supports a wide range of inputs and remains compatible with other conditionals, including text, fine-structure (e.g.\ edge maps), and style.

While effective, Colorful-Noise has several limitations. Beyond those discussed in \cref{subsec: ablation}, it requires careful tuning of $\alpha$ and $\gamma$ based on the frequency characteristics of the conditional input. Some models, such as Flux, are more sensitive to noise manipulation, requiring lower values for stability and limiting the use of realistic images for guidance. As a result, the method is more interactive and less suited for fully automated large-scale generation. Additionally, because conditioning is applied in latent space, masks are significantly downsampled, resulting in coarse control and limiting fine-grained guidance. We discuss this limitation further in the supplementary material. Future work could explore learning the relationship between images and noise latents to enable automatic parameter selection across varying inputs and distributions.

While we demonstrate this approach on images, extending the analysis of frequency components to other domains such as video and 3D may reveal additional benefits. Further investigation could also improve controllability, including high-frequency conditioning and disentangling color and structural features. We leave these directions for future work.



\section*{Acknowledgments}
This work was partially supported by Israel Science Foundation Grant no. 1427/25 and 
Joint NSFC-ISF Research Grant no. 3077/23.

\bibliographystyle{ACM-Reference-Format}
\bibliography{main}

\clearpage

\appendix
\begin{center}
    {\LARGE \textbf{Supplemental Material}}\\[1em]
\end{center}

\vspace{2em}





\settopmatter{printacmref=false}
\renewcommand\footnotetextcopyrightpermission[1]{}






\title{Colorful-Noise: Low-Frequency Noise Manipulation for Color-Based Conditional Image Generation - Supplemental Material}

\section{Appendix A. ---  Analysis}
\label{sec:Analysis}
We share additional information from our analysis.

\subsection{Latent Frequency Bands Influence on Generation}
Subsection 3.1 in the main manuscript presents a t-SNE visualization of the generated images based on their LPIPS distance. The visualization presents the image sharing low-frequencies by color and and example of two clusters colored by their mid frequencies. To get a broader picture of the spread of all frequency groups we share the plots with all frequency group separations in \cref{fig:frequency_mix_supp}. Additionally, we conduct the experiment on 40 prompts and quantitatively evaluate the average clustering quality of images grouped by each frequency band. We use a silhouette score computed with LPIPS distances to measure perceptual clustering quality. We consider two settings: (1) the clustering score of each frequency over the entire generated evaluation set (first-order), and (2) the clustering score within clusters defined by another frequency (second-order). The results are shown in \cref{tab:exp1}. As observed, across the full evaluation set, images sharing the same low-frequency components achieve the highest silhouette score, consistent with the left plot in \cref{fig:frequency_mix_supp}. In the second-order analysis, mid frequencies exhibit high silhouette scores within low-frequency clusters, again aligning with the middle plot in \cref{fig:frequency_mix_supp}. High frequencies do not form strong clusters in either setting, which also agrees with the trends shown in \cref{fig:frequency_mix_supp}.

\begin{table}[H]
\centering
\small
\setlength{\tabcolsep}{6pt}
\caption{\textbf{First- and second-order clustering across frequency bands.}
First-order scores measure clustering over the entire evaluation set. 
Second-order scores measure clustering within clusters defined by the row frequency. 
Low frequencies cluster best globally, while mid frequencies cluster strongly within low-frequency clusters. 
High frequencies do not exhibit meaningful clustering in either setting.}
\label{tab:frequency_clustering}
\begin{tabular}{lccc}
\toprule
 & Low & Mid & High \\
\midrule
\textbf{First-order (Global)} 
& \textbf{0.375} & 0.0311 & -0.011 \\
\midrule
\textbf{Second-order (Within clusters)} \\
Within low-frequency clusters 
& -- & \textbf{0.4284} & -0.079 \\
Within mid-frequency clusters 
& \textbf{0.6655} & -- & -0.0783 \\
Within high-frequency clusters 
& \textbf{0.3514} & -0.0325 & -- \\
\bottomrule
\end{tabular}
\label{tab:exp1}
\end{table}

\begin{figure}[t]
    \centering
    \includegraphics[width=\columnwidth]{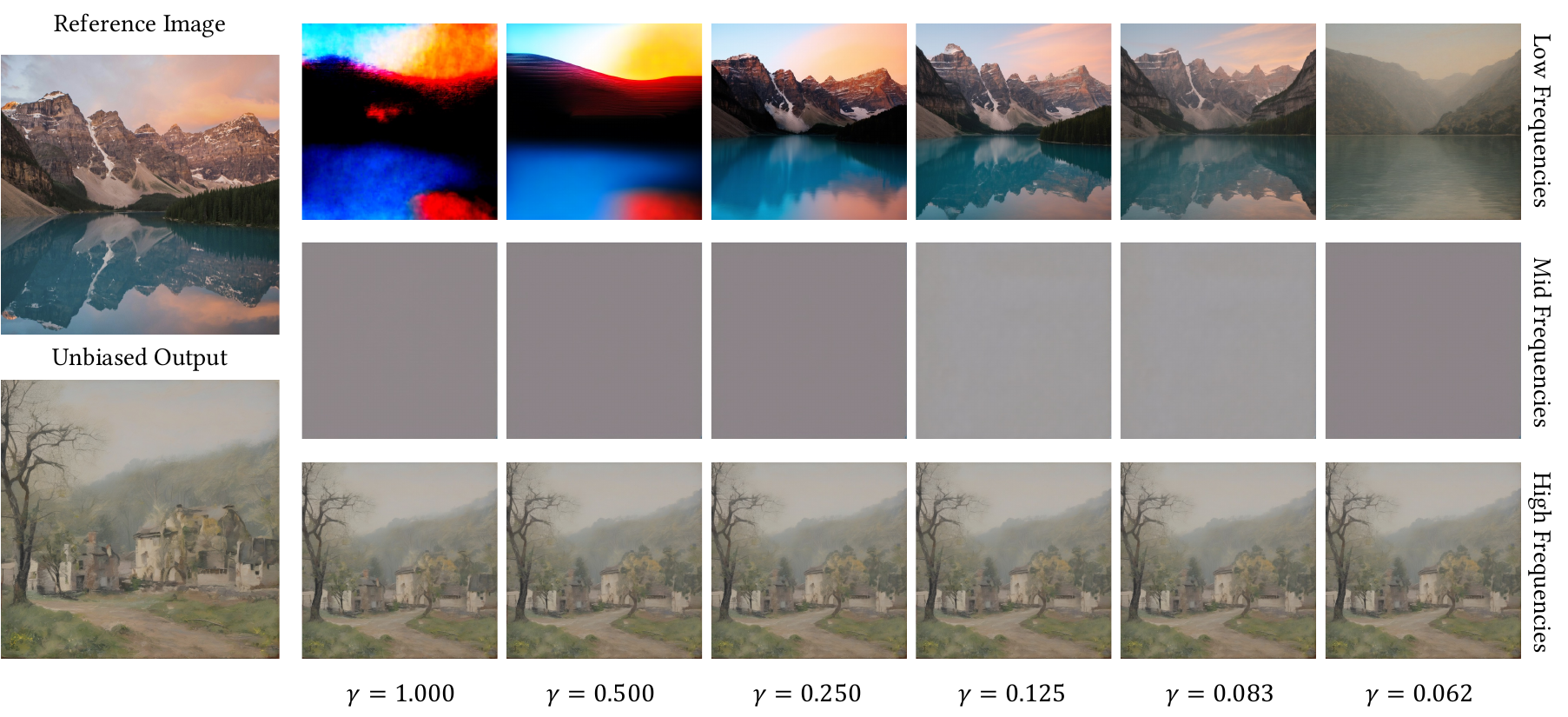}
    \caption{\textbf{Noise and Image Combinations.} Images generated using noise injected with Mid-Frequencies cause the diffusion process to collapse and images generated using noise injected with High-Frequencies show minimal impact on the output, apparent by the similarity of the result to the unbiased output.}
    \label{fig:analysis_injection}
\end{figure}

\begin{figure*}[t]
    \centering
    \includegraphics[width=\textwidth]{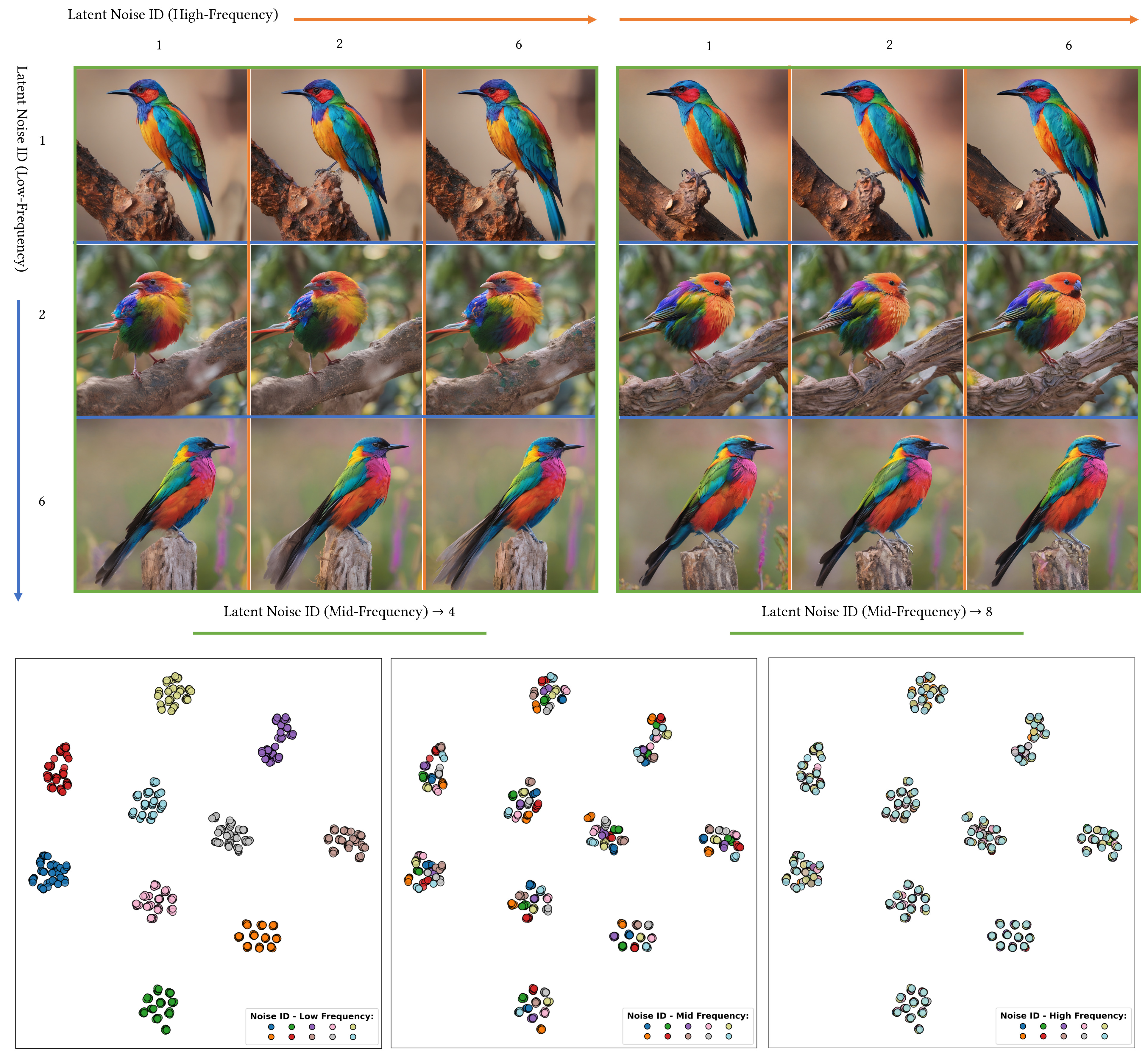}
    \caption{\textbf{Frequency Mixing in Noise Latents.} Additional example and full t-SNE plots with labels for low-, mid-, and high-frequencies.}
    \label{fig:frequency_mix_supp}
\end{figure*}

\begin{table}[t]
\centering
\footnotesize
\setlength{\tabcolsep}{3.5pt}
\caption{\textbf{Compositing natural frequencies into noise.} 
We inject low-, mid-, or high-frequency components from a reference image into the input noise. 
We report LPIPS distance to the reference image and frequency-wise cosine similarity between generated and reference images. 
Low-frequency injection yields the strongest perceptual similarity, while mid- and high-frequency injection fail to preserve the corresponding frequency content.}
\label{tab:frequency_injection}
\begin{tabular}{lcc}
\toprule
 & LPIPS $\downarrow$ & Freq. Similarity ($cos)\uparrow$  \\
\midrule
Low-freq injection. & \textbf{0.546} & \textbf{0.947} \\
Mid-freq injection. & 0.888 & 0.148 \\
High-freq injection. & 0.705 & 0.012 \\
\bottomrule
\end{tabular}
\label{tab:injection_tab}
\end{table}

\subsection{Compositing Natural Frequencies into Noise}

In the main manuscript, we explore compositing low-frequency components from a natural image into the input noise, motivated by our observation that low frequencies dominate white Gaussian noise. In this subsection, we extend this analysis to mid- and high-frequency bands. We further repeat the experiment on 500 images from the Aesthetic-4K dataset~\cite{zhang2025diffusion4k}, and report quantitative results using two metrics: (1) LPIPS distance~\cite{lpips} to the reference image, and (2) frequency-wise cosine similarity (low, mid, and high) between the reference and generated images. For each example, we generate outputs using $\gamma \in {1, 0.5, 0.25, 0.125, 0.083, 0.062}$ and report the best-performing result for each sub-band.

\begin{figure*}[t]
    \centering
    \includegraphics[width=\textwidth]{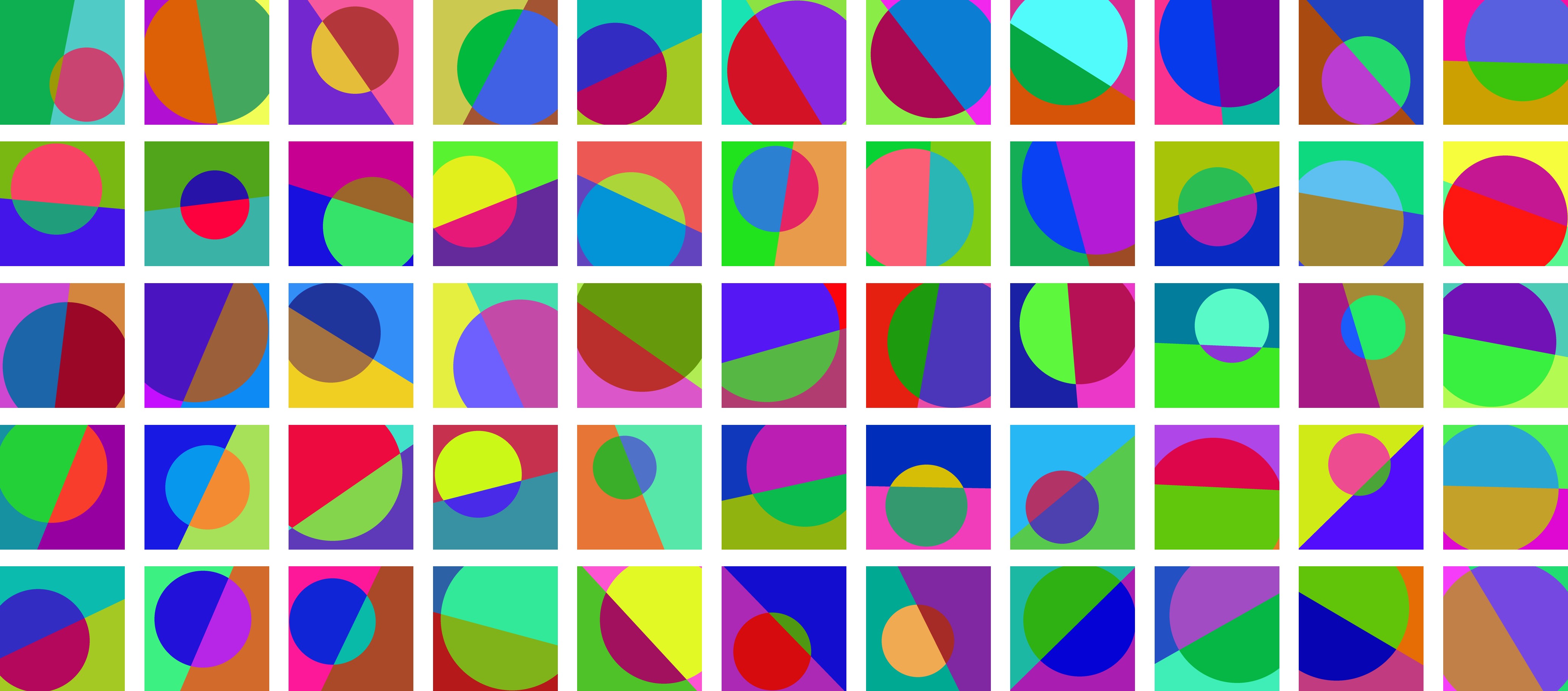}
    \caption{\textbf{Color Evaluation Set.} A small set of examples from the colorful evaluation dataset, highlighting its variability.}
    \label{fig:color_examples}
\end{figure*}

We visualize the results in \cref{fig:analysis_injection}, using the same reference image as in Sec.~3.3 of the main manuscript for consistency. As shown, injecting low-frequency components produces an image that closely resembles the reference. In contrast, mid-frequency injection results in a near-uniform gray image, indicating a collapse of the diffusion process, while high-frequency injection has only a minor effect, with outputs remaining similar to unconditioned samples. Quantitative results are reported in \cref{tab:injection_tab}. Consistent with the qualitative observations, low-frequency injection achieves the best LPIPS score, reflecting the highest perceptual similarity to the reference. Moreover, although low-frequency content is partially preserved, mid- and high-frequency components are not retained under injection, as indicated by the cosine similarity measured within each frequency band.

\subsection{Evaluation Dataset}
In \cref{fig:color_examples}, we use a synthetic dataset to evaluate the reconstruction and manipulation capabilities of the generation process under injected frequencies. The dataset was specifically designed to assess the reconstruction of geometric shapes with varying colors. For each image, we randomly sample the size and center of a circle, the position and orientation of a line, and assign a random color to each intersection region. Since our focus is on low-frequency information, the dataset consists of images dominated almost entirely by low-frequency components. Additional examples from the dataset are provided in \cref{fig:color_examples}.

\section{Appendix B. ---  Method}
\label{sec:appendix_b_analysis}
In the main manuscript, we present our method using the Discrete Fourier Transform (DFT) to decompose a signal into its frequency components. In recent years, several diffusion-based works~\cite{kim2024diffusehigh, hiWave} have instead adopted the Discrete Wavelet Transform (DWT) to improve image-focused methods. In the appendix, we explore an alternative formulation based on colorful noise with DWT rather than DFT. We first formalize this approach and then present comparative results. 

\subsection{Wavelet-Based Colorful-Noise}
\label{subsec:supp_wavelet_formulation}
Wavelets are a family of functions which receive an image and decomposes it to 4 components: LL, LH, HL, HH, where LL represents the low frequencies of an image and LH, HL, and HH are considered to hold its high frequencies. The function also receives a parameter J which controls the level of decomposition, where for $J=i$ the function runs recursively on the previous low level frequencies $LL_{(i-1)}$, where $LL_0$ is the input image. Formally: Let $\mathcal{W}$ be a wavelet function, then for an input Image $I$ and decomposition level $J=i > 0$ we have:
\begin{equation}
    (LL, LH, HL, HH)_i = \mathcal{W}(LL_{(i-1)})
    \qquad
    LL_0 = I
\end{equation}

To reconstruct the image a series of recursive inverse functions are applied to reconstruct each LL image all the way to I:

\begin{equation}
    LL_i = \mathcal{W}^{-1}([LL, LH, HL, HH]_{(i+1)})
\end{equation}

For simplicity of read, we consider $H$ to be the collection of all high frequencies: $H = [LH, HL, HH]$.

\subsection{Low-Frequency Color Conditioning}
Let $C$ be a conditioning image, and let $z \sim \mathcal{N}(\mathbf{0}, \mathbf{I})$. Since the denoising process of $z$ occurs in the VAE latent space, we first encode $C$ into this space:

\begin{equation}
c = \mathcal{E}(C).
\end{equation}

Then, to condition $z$ with $c$ we decompose both latents using a wavelet decomposition function for some $J=i>0$:

\begin{equation}
[LL, H]_c = \mathcal{W}(c)
\qquad
[LL, H]_{z} = \mathcal{W}(z)
\end{equation}

to condition the latent noise $z$ with the latent condition $c$ we simply compose the low-frequencies of $x$ with the high frequencies of $\mathbf{z}$ using the inverse wavelet function:

\begin{figure*}[t]
    \centering
    \includegraphics[width=\textwidth]{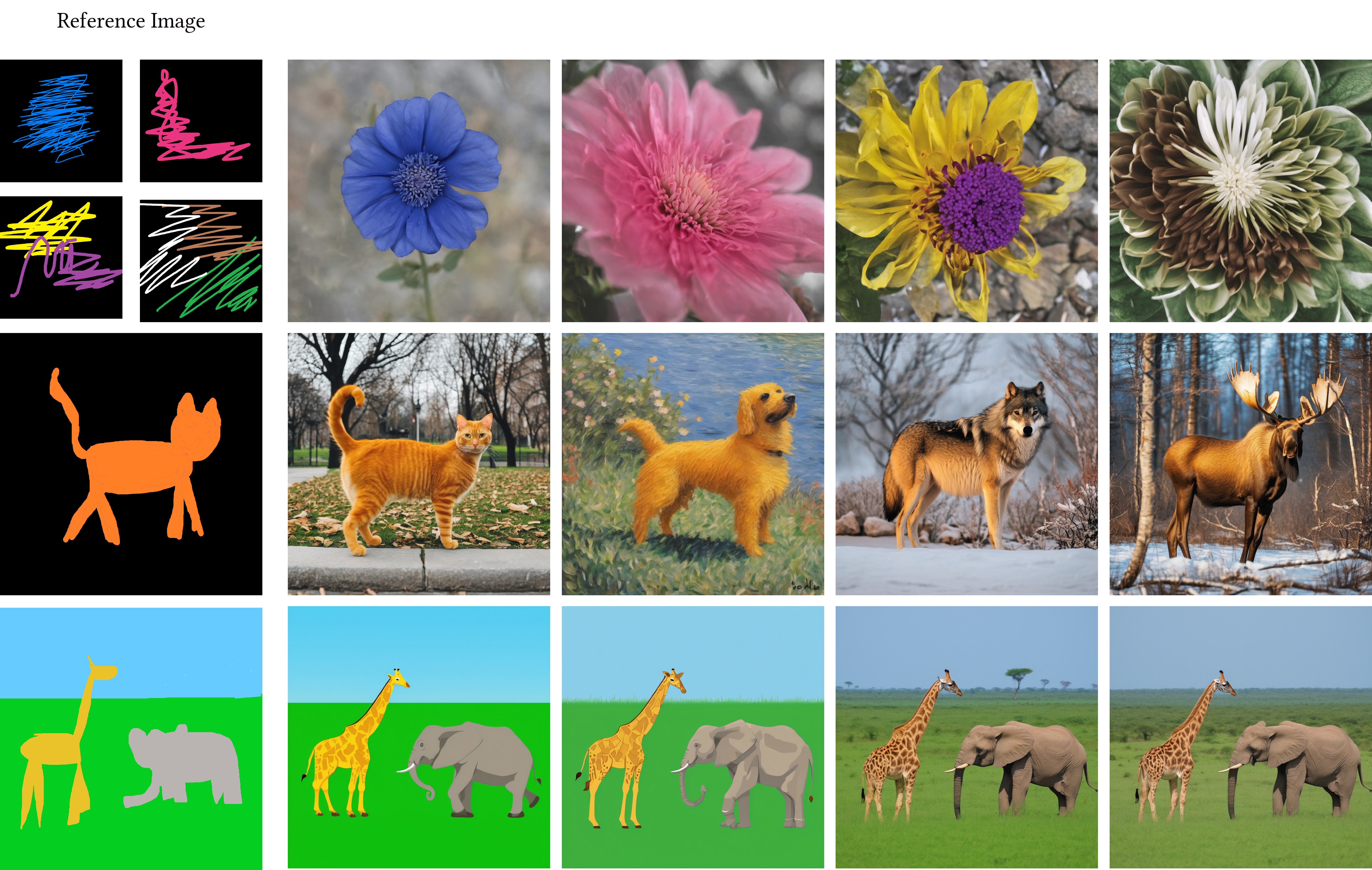}
    \caption{\textbf{Wavelet Colorful-Sketch Conditioning Results.} Results of applying colorful noise using wavelets for frequency decomposition. We present results with identical conditional inputs to enable direct comparison.}
    \label{fig:wavelets_results}
\end{figure*}

\begin{figure*}[t]
    \centering
    \includegraphics[width=\textwidth]{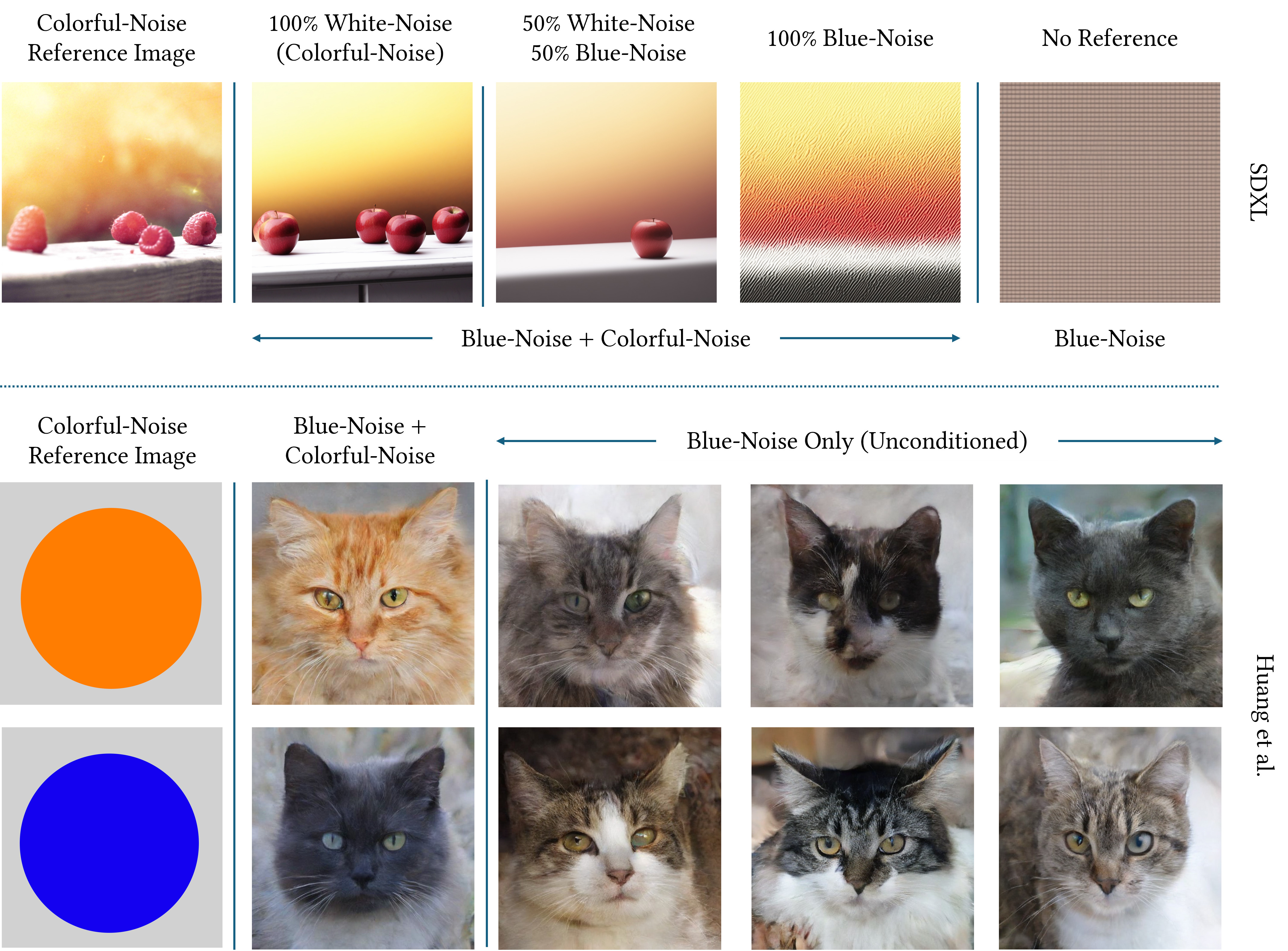}
    \caption{\textbf{Colorful Noise for Blue-Noise Diffusion.} We evaluate blue-noise inputs with SDXL (top) and a blue-noise–trained model from Huang et al. (bottom). For both, we compare blue-noise generation with and without colorful-noise conditioning. Interpolating from white to blue noise in SDXL, as proposed by Huang et al., leads to degraded results; however, adding colorful-noise conditioning improves outputs compared to pure blue noise alone. The unconditional blue-noise model of Huang et al. produces cats with randomly varying colors, whereas simple colorful-noise inputs effectively guide the model to generate cats with controlled colors, such as orange and bluish tones.}
    \label{fig:blue-noise}
\end{figure*}

\begin{equation}
z^{c} = \mathcal{W}^{-1}(\gamma * LL_c, H_z)
\end{equation}

where $\gamma\in\mathbb{R}$ serves as a scaling factor.

\subsection{Wavelet Results}
\label{subsec:supp_wavelet_results}
We present a sample of results in \cref{fig:wavelets_results} for sketch-based conditioning and in \cref{fig:ADDITIONAL_SDXL_WAVELETS} for color–style alignment using SDXL. Additionally, we add FFT based results for color-based style alignment in \cref{fig:wavelets_examples_2} we use the same conditional inputs as in the main manuscript for easy comparison. We set $J=3$ and $\gamma=0.083$ for sketch-based conditioning, and $J=3$ and $\gamma=0.2$ for color–style alignment. As observed, the results are qualitatively comparable to those obtained using DFT-based decomposition. Nevertheless, we adopt FFT in the main method, as it achieves similar performance while providing a more fine-grained frequency decomposition, which proved beneficial for our analysis and experiments.

\begin{figure*}
    \centering
    \includegraphics[width=.86\textwidth]{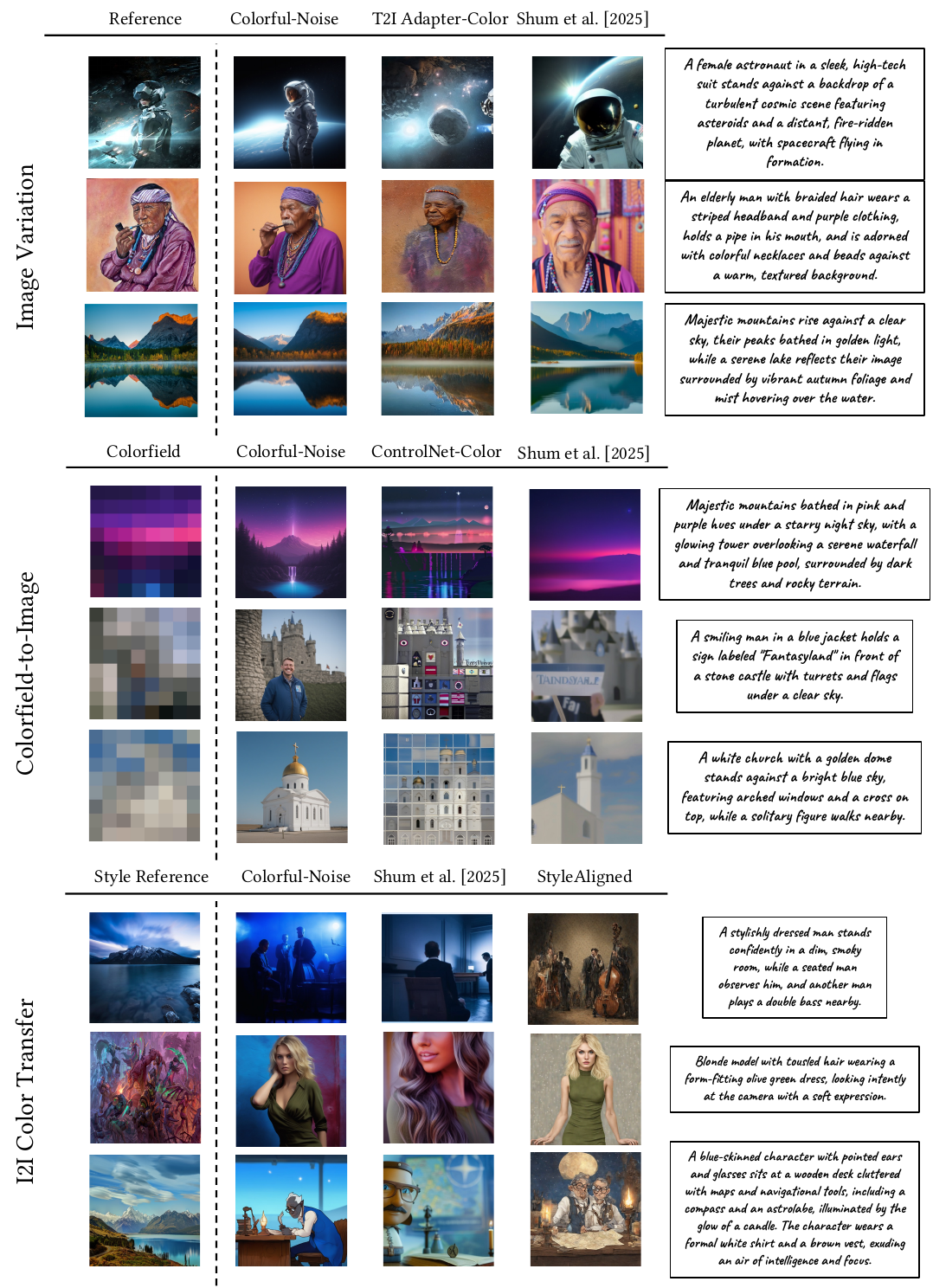}
    \vspace{-1em}
    \caption{\textbf{Qualitative Comparison to Previous Works.} ColorfulNoise supports multiple tasks, including Image Variation generation, Colorfield-to-Image synthesis and Image Color Transfer. Shown are representative results on the Aesthetic-4K evaluation set.}
    \label{fig:comparisons}
\end{figure*}

\begin{figure*}[t]
    \centering
    \includegraphics[width=\textwidth]{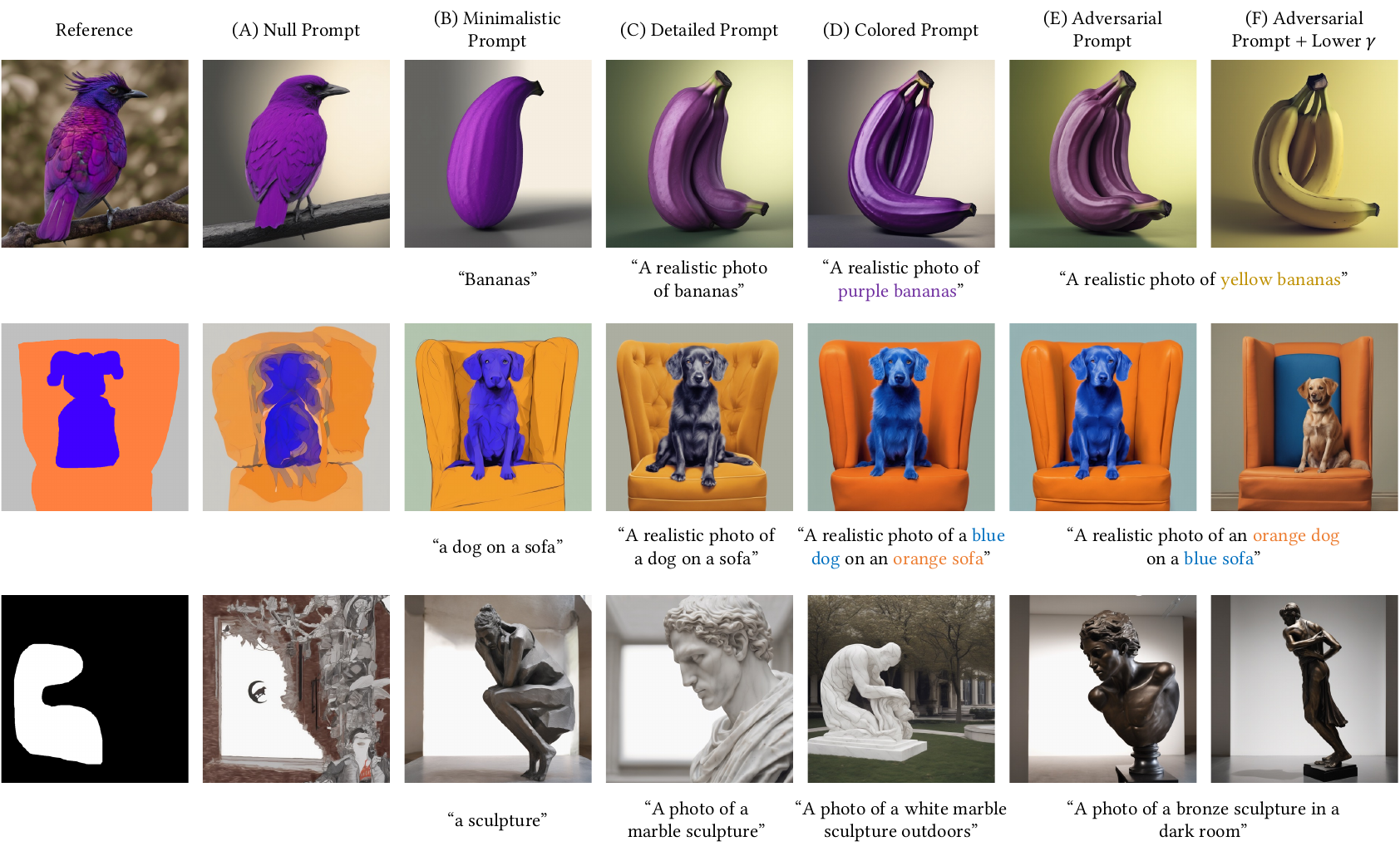}
    \caption{\textbf{Prompt Ablation Study.} We investigate the effect of various prompt levels on the output, when conditioned with a photo, a full colorful-sketch, and a masked colorful-sketch. As there is no semantic matching between colors and subject in the prompt, the generation model is free to interpret the color mapping freely which leads to various semantic interpretations based on the input prompt.}
    \label{fig:frequency_edit}
\end{figure*}

\begin{figure*}[t]
    \centering
    \includegraphics[width=\textwidth]{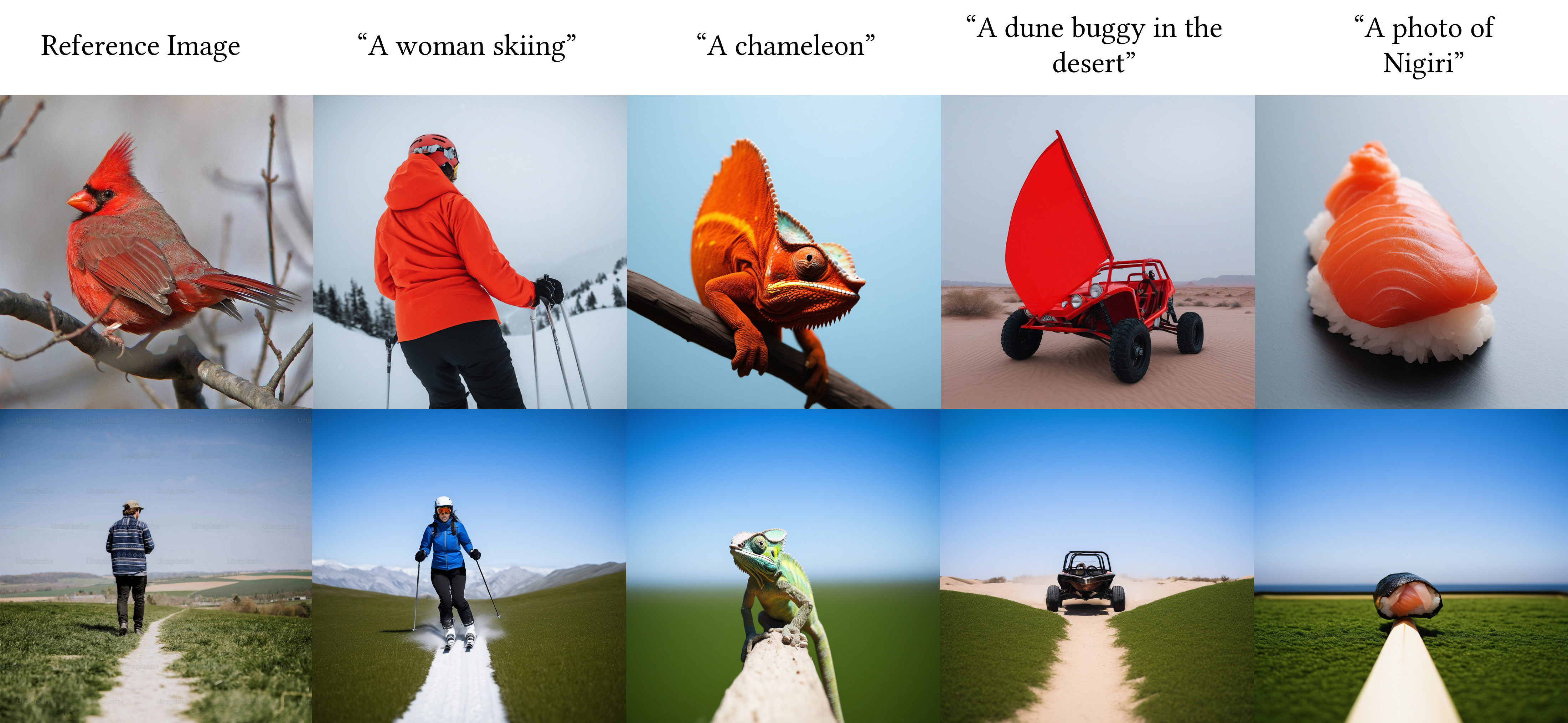}
    \caption{\textbf{Additional Results.} Additional results for Color-Based Style Alignment with SDXL.}
    \label{fig:ADDITIONAL_SDXL_WAVELETS}
\end{figure*}

\begin{figure*}[t]
    \centering
    \includegraphics[width=\textwidth]{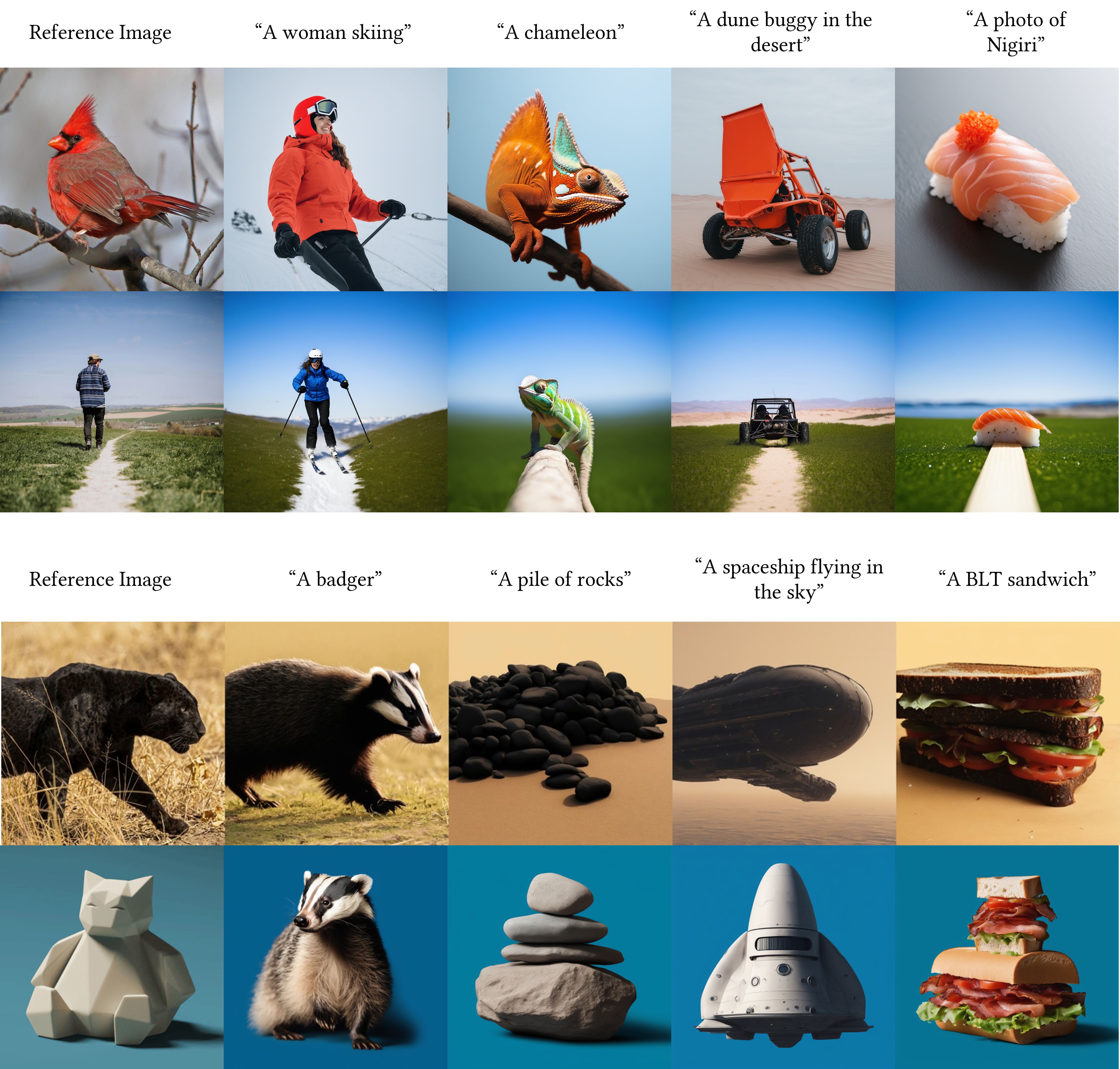}
    \caption{\textbf{Wavelets Color-Based Style Alignment Results.} Results of applying colorful noise via wavelet-based frequency decomposition. All results use the same conditional inputs to enable direct comparison. Bottom reference © Augustin Arroyo (@flowalistic on Instagram). All rights reserved.}
    \label{fig:wavelets_examples_2}
\end{figure*}

\section{Appendix C. ---  Applications}
\label{sec:appendix_d_sd35}


\subsection{Additional Results}
\begin{figure*}[t]
    \centering
    \includegraphics[width=\textwidth]{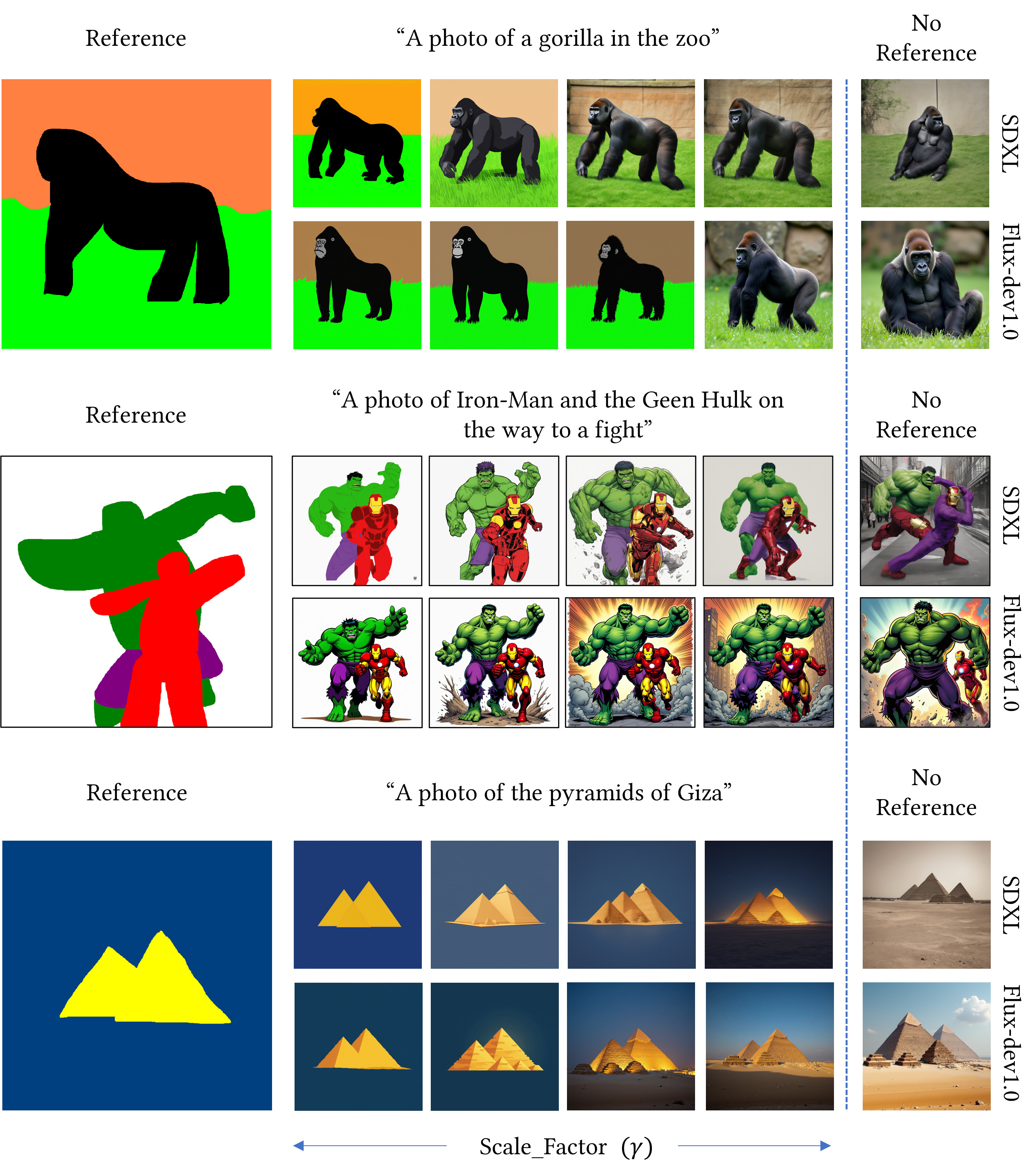}
    \caption{\textbf{Additional Results.} Additional results illustrating variations induced by changes in the $\gamma$ scale-factor.}
    \label{fig:interp_results_gamma}
\end{figure*}

\begin{figure*}[t]
    \centering
    \includegraphics[width=\textwidth]{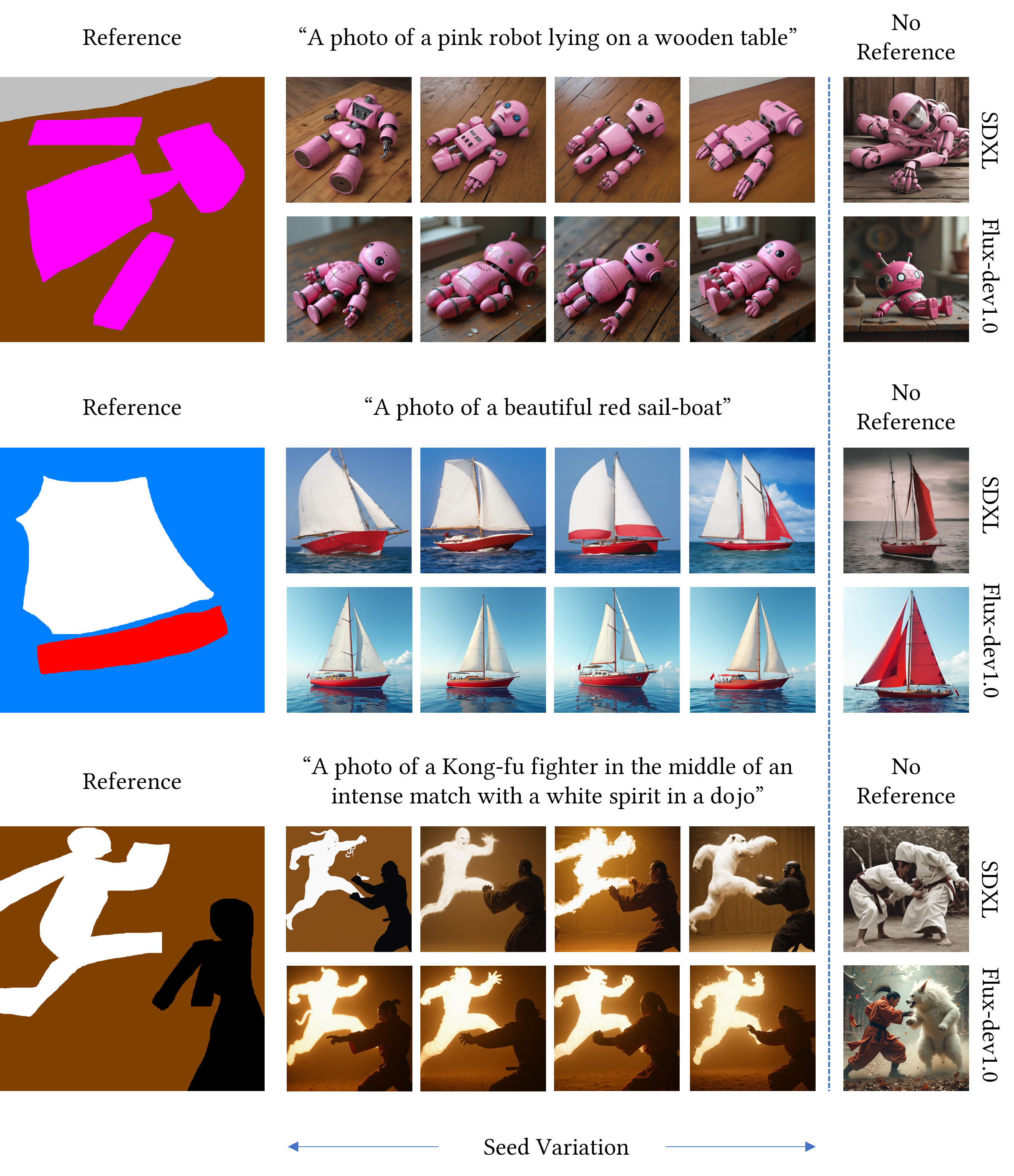}
    \caption{\textbf{Additional Results.} Additional results illustrating variations induced by changes in the random seed.}
    \label{fig:interp_results_seed}
\end{figure*}

\begin{figure*}[t]
    \centering
    \includegraphics[width=\textwidth]{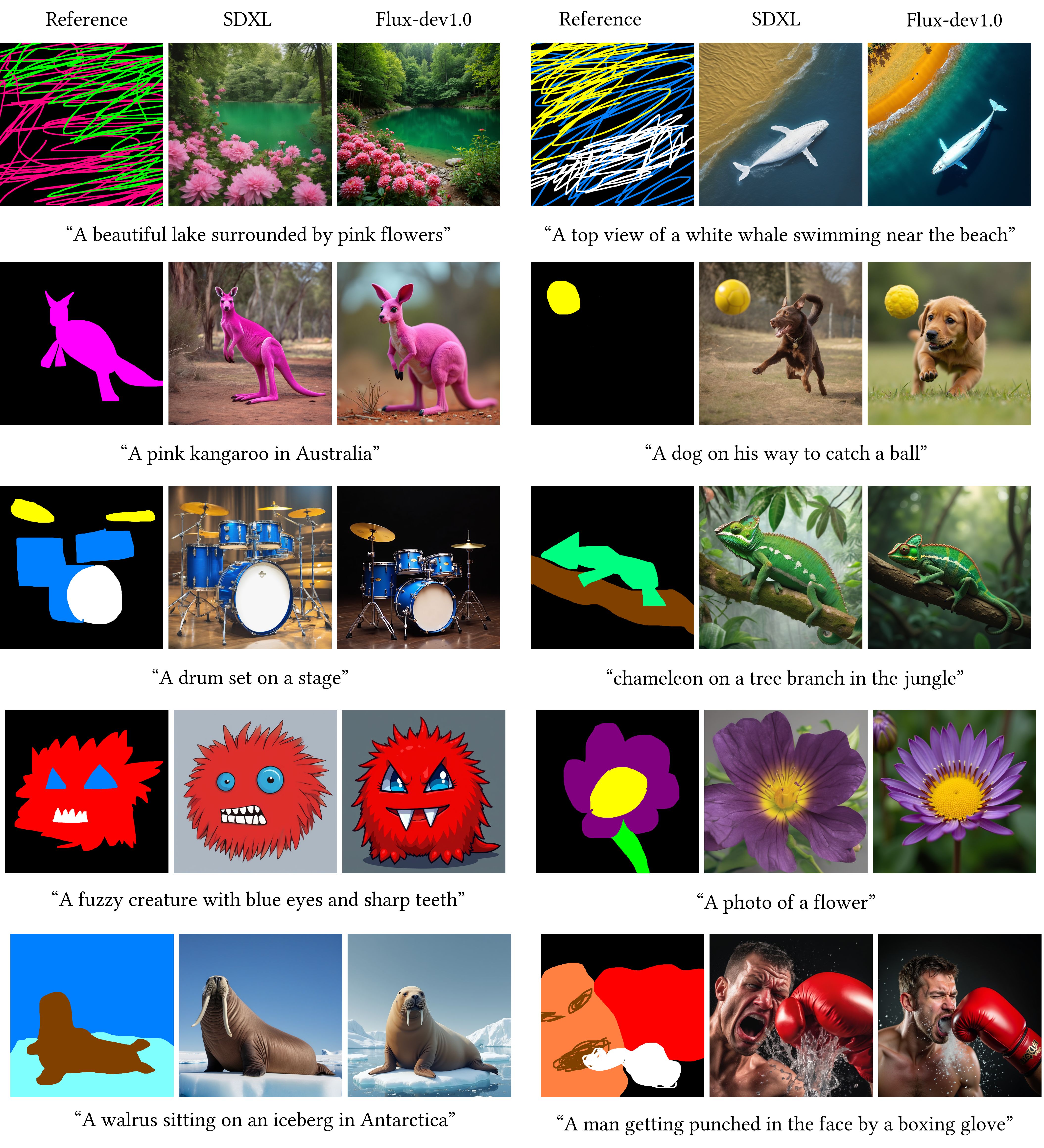}
    \caption{\textbf{Additional Results.} Additional results for masked-sketches (Top 4 rows) and full-sketches (last row).}
    \label{fig:masked_results_1}
\end{figure*}

\begin{figure*}[t]
    \centering
    \includegraphics[width=\textwidth]{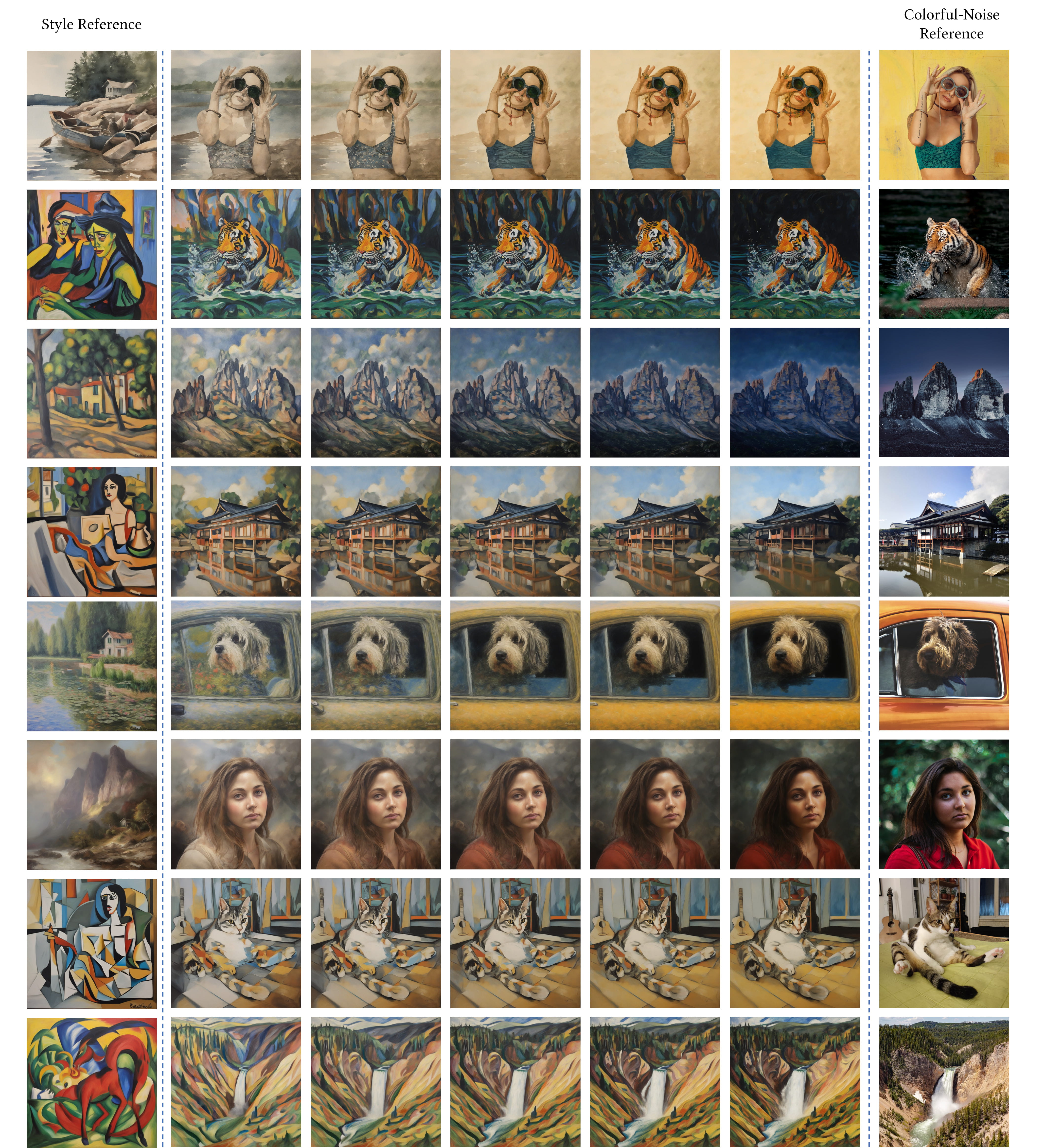}
    \caption{\textbf{Additional Results.} Additional results for Color-Preserving Stylization. }
    \label{fig:masked_results_2}
\end{figure*}

\begin{figure*}[t]
    \centering
    \includegraphics[width=\textwidth]{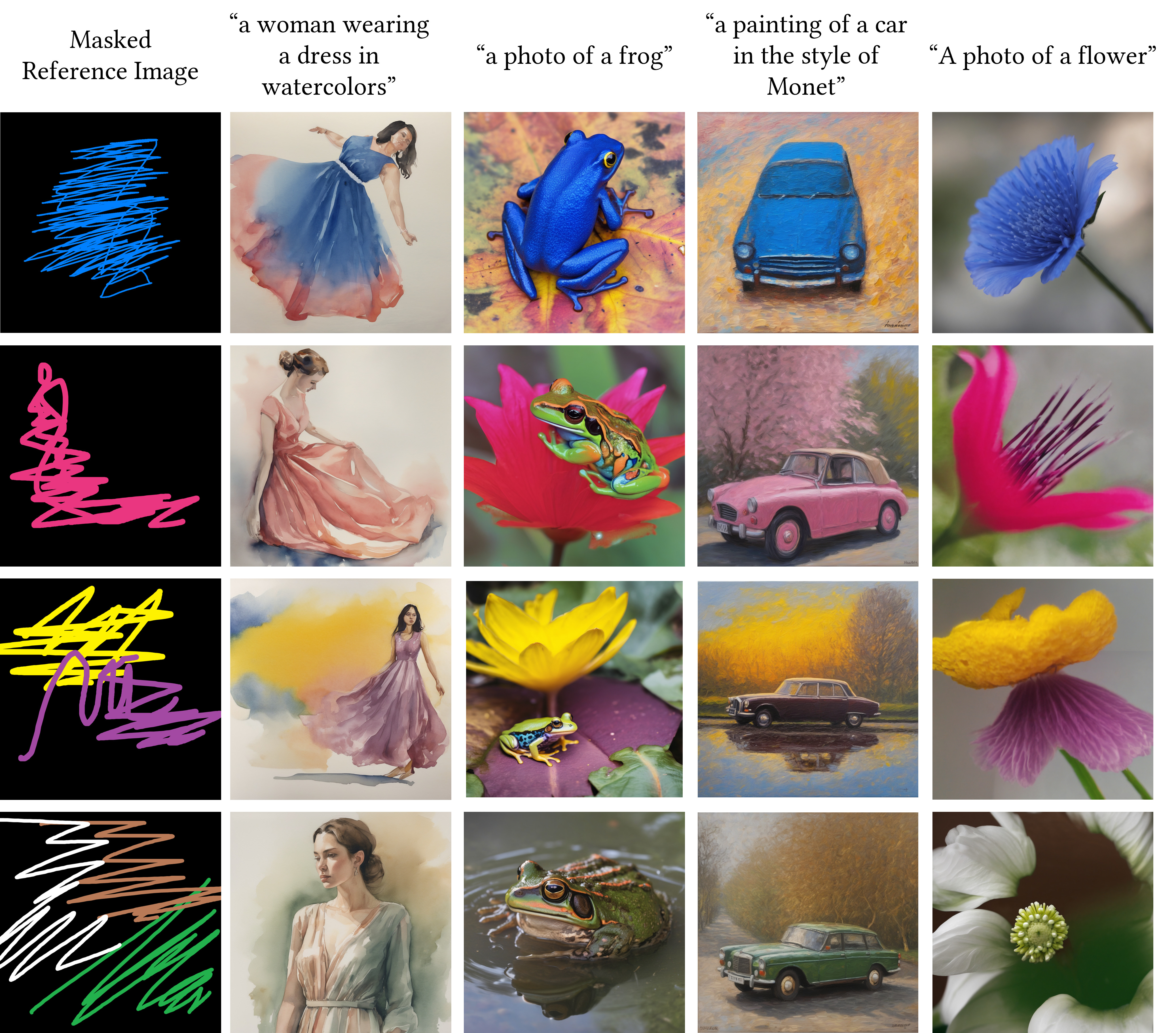}
    \caption{\textbf{High-Resolution Results.} High resolution results from Fig. 6 of the main manuscript. }
    \label{fig:res0}
\end{figure*}

\begin{figure*}[t]
    \centering
    \includegraphics[width=\textwidth]{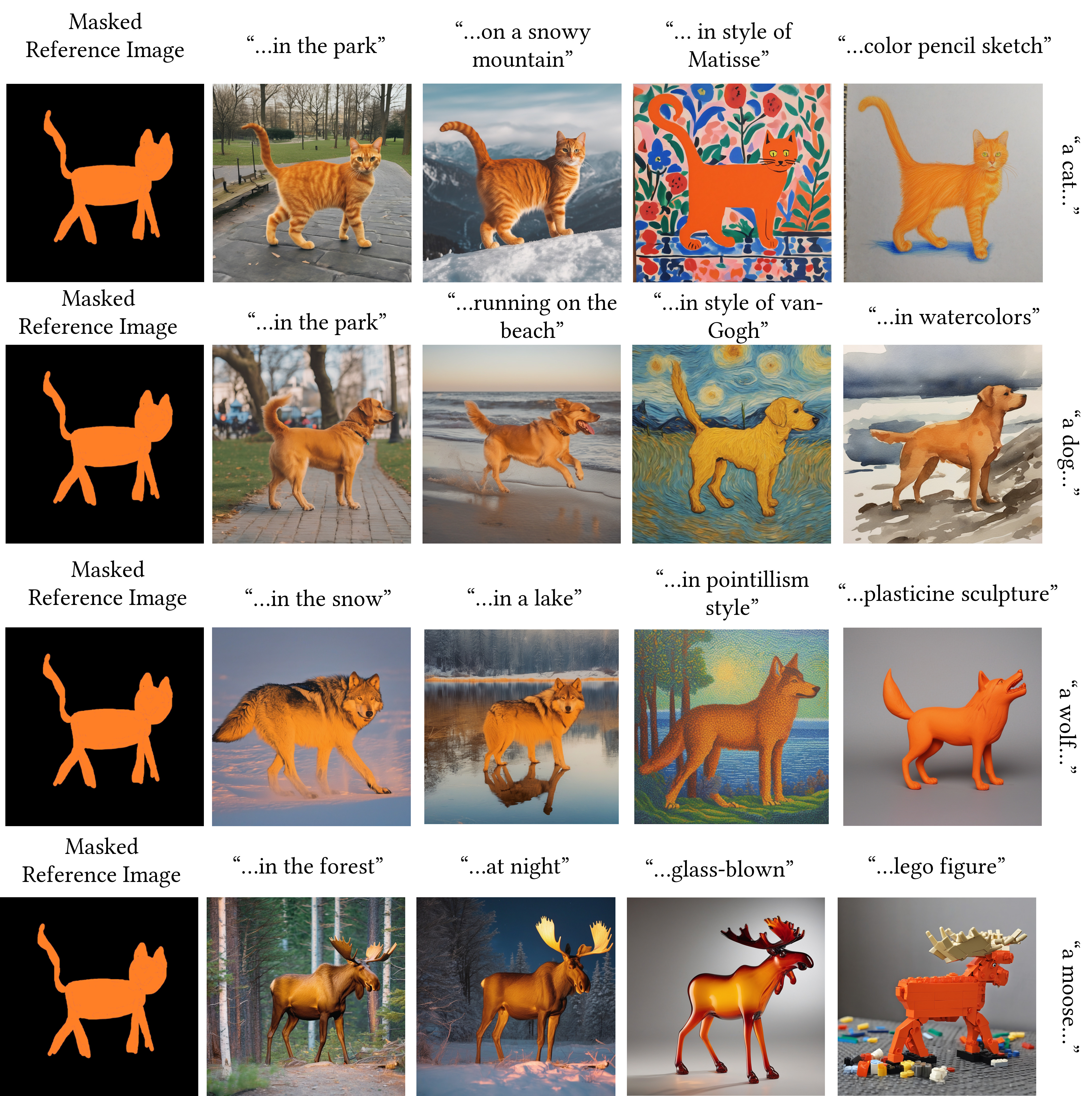}
    \caption{\textbf{High-Resolution Results.} High resolution results from Fig. 6 of the main manuscript. }
    \label{fig:res1}
\end{figure*}

\begin{figure*}[t]
    \centering
    \includegraphics[width=\textwidth]{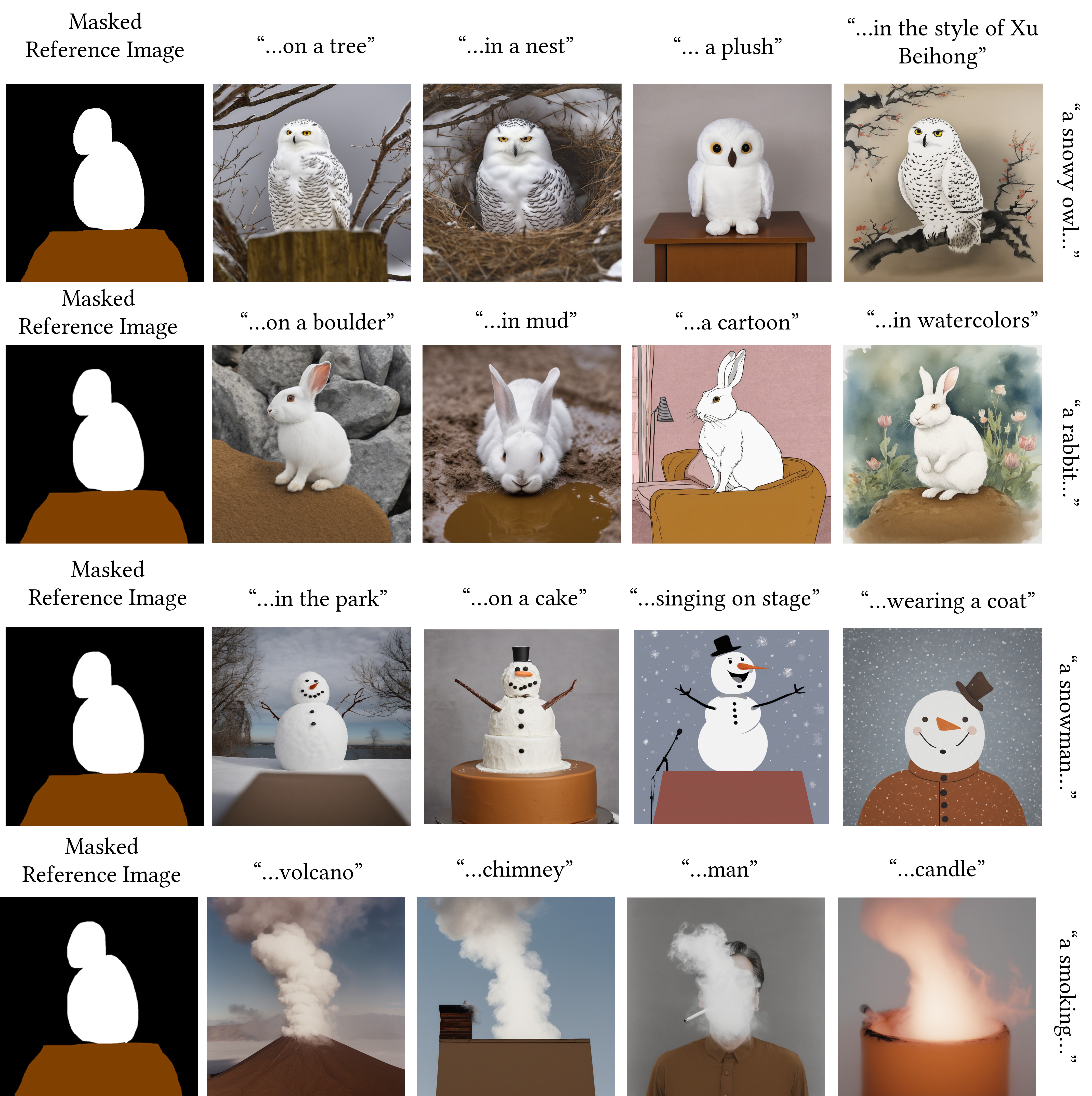}
    \caption{\textbf{High-Resolution Results.} High resolution results from Fig. 6 of the main manuscript. }
    \label{fig:res2}
\end{figure*}

\begin{figure*}[t]
    \centering
    \includegraphics[width=\textwidth]{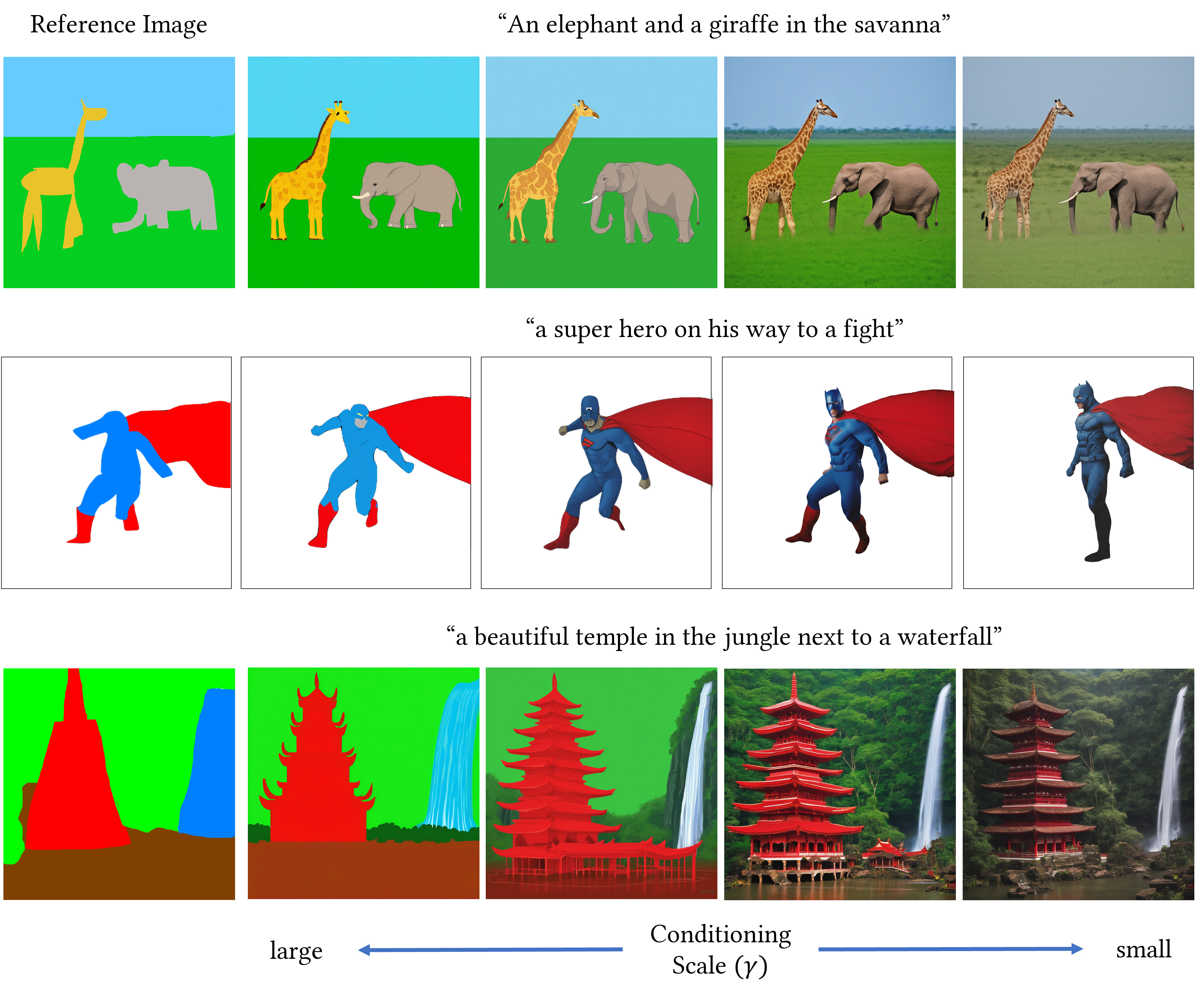}
    \caption{\textbf{High-Resolution Results.} High resolution results from Fig. 6 of the main manuscript. }
    \label{fig:res3}
\end{figure*}

We present additional results for both SDXL~\citep{podell2023sdxlimprovinglatentdiffusion} and Flux-dev1.0~\citep{flux2024}. \cref{fig:interp_results_gamma} illustrates examples of interpolating different $\gamma$ values, \cref{fig:interp_results_seed} shows variations across different random seeds for the same prompt and color conditioning, and \cref{fig:masked_results_1} presents results using both masked and full sketch conditional inputs. Additionally, we provide additional results in~\cref{fig:interp_results_gamma,fig:interp_results_seed,fig:masked_results_1,fig:masked_results_2,fig:res0,fig:res1,fig:res2,fig:res3}

\section{Appendix E --- Blue-Noise}
Prior work~\cite{blue_noise_in_diffusion, inverse_heat, nonisotropic} explores the use of non-white noise in diffusion models. In particular,~\cite{blue_noise_in_diffusion} argues that blue noise~\cite{blue_noise}, which has no energy in its low-frequencies, may be better suited for early diffusion timesteps, as it introduces additional structure in the initial stages, aligning with the frequency progression of diffusion models. Our findings support this claim, showing that structured signals can replace white noise in the low-frequency components. Unlike Huang et al., we extract low-frequency components from images to explicitly condition the output on specific structures. Motivated by the connection to our findings, we conduct experiments that compare and combine blue-noise with colorful-noise.

\subsection{Blue-Noise for White-Noise Diffusion}
We begin by asking a natural question: can blue-noise replace white-noise in diffusion models that were pretrained using white-noise? To evaluate this, we follow Huang et al.~\cite{blue_noise_in_diffusion} and use blue-noise as input to SDXL~\cite{podell2023sdxlimprovinglatentdiffusion}, which was originally trained on white Gaussian noise. We further examine whether blue-noise can substitute white Gaussian noise when combined with colorful-noise, i.e., when the low-frequency components of blue-noise are replaced with those from images. Results are shown in \cref{fig:blue-noise} (Top). As observed, unlike colorful-noise, blue-noise alone cannot effectively replace white-noise when the model was not specifically trained for it. Nevertheless, replacing the low-frequency components of blue-noise with those from images leads to improved results, suggesting that colorful-noise brings blue-noise closer to a distribution that is compatible with the model.

\subsection{Colorful-Noise for Blue-Noise Diffusion}
In~\cite{blue_noise_in_diffusion}, Huang et al. train unconditioned diffusion models using blue-noise inputs. Since these models are not text-conditioned, each model is trained for a specific object class and generates random samples of that class without explicit control over the output beyond iterative generation. To assess the extensibility of colorful-noise, we apply it to blue-noise inputs and use the resulting signals for conditional generation with the model trained by Huang et al. Results are presented in \cref{fig:blue-noise} (Bottom). As shown, when applying colorful-noise to blue-noise inputs in a cat-specific model, we are able to influence the color of the generated cat. This demonstrates that colorful-noise is not only extensible to other diffusion models, but can also be applied to alternative noise distributions such as blue-noise, enabling conditional image generation even in models that lack text conditioning.

\section{Appendix D. ---  Qualitative Results}
\subsection{Comparison with Prior Works}
To further illustrate the effectiveness of our method, \Cref{fig:comparisons} presents qualitative comparisons with prior works~\cite{Shum_2025_CVPR,10.1609/aaai.v38i5.28226,Hertz_2024_CVPR}. For each task, we select the most relevant baselines and visualize examples from the Aesthetics-4K dataset~\citep{zhang2025diffusion4k}. The figure demonstrates that Colorful-Noise produces color-preserving results across different guidance types while maintaining alignment with the textual prompts.

\subsection{Ablation Study}
In \cref{fig:frequency_edit} we provide a higher restitution version for our prompt ablation study (discussed in the main paper).

\section{Appendix F. ---  Limitations}
In the main manuscript, we discuss a key limitation arising from downscaling the guidance input to the VAE latent space. This process restricts our method’s ability to handle fine-grained details in the input, which may disappear entirely in the worst case or become sufficiently small for the model to ignore. This highlights a fundamental trade-off between training-free and training-based approaches.

Training-free, intervention-free methods such as Colorful Noise are not restricted to a single input modality, making them flexible and broadly applicable. However, this flexibility also grants the model greater freedom in interpreting the conditioning signal—allowing creative and robust generations, as demonstrated in our examples, but also enabling the model to ignore fine or ambiguous details. In contrast, training-based methods are optimized for specific inputs and therefore adhere more strictly to the conditioning signal. While this reduces flexibility, it results in more reliable and precise outputs, making such methods better suited for large-scale generation and ease of use.

The effects of guidance downscaling are evident in \cref{fig:interp_results_gamma,fig:interp_results_seed,fig:masked_results_1}, where the generated outputs do not always exactly match the input masks. For example, in the robot example of \cref{fig:interp_results_seed}, despite a clearly square head in the input, the model often produces a rounded humanoid head, reflecting a bias toward its learned distribution. Similarly, in the second row of \cref{fig:interp_results_gamma}, the generated pose deviates from the precise pose specified in the guidance map.

While this behavior limits precision, we find it advantageous in scenarios where only approximate correspondence is desired. This aligns with the goal of our method, which enables sketch-based inputs without requiring professional-level accuracy. Future work may explore mechanisms for amplifying subtle details to improve exact adherence to the input mask.

\end{document}